%% file: cas-sc-template.tex
\documentclass[a4paper,fleqn]{cas-sc}
\usepackage[authoryear]{natbib}
\usepackage[vietnamese,english]{babel}
\usepackage[T5,T1]{fontenc}
\usepackage{todonotes}
\usepackage{subcaption}
\def\tsc#1{\csdef{#1}{\textsc{\lowercase{#1}}\xspace}}
\tsc{WGM}
\tsc{QE}
\tsc{EP}
\tsc{PMS}
\tsc{BEC}
\tsc{DE}

\newcommand*\vn{\fontencoding{T5}\selectfont\selectlanguage{vietnamese}}
\newcommand*\en{\fontencoding{T1}\selectfont\selectlanguage{english}}
\newcommand{\sstitle}[1]{\smallskip\noindent\textbf{#1.\/}}

\begin{document}
\let\WriteBookmarks\relax
\def\floatpagepagefraction{1}
\def\textpagefraction{.001}

\shorttitle{Vietnamese Visual Reasoning }

\shortauthors{Khiem Vinh Tran et~al.}

\title [mode = title]{ViCLEVR: A Visual Reasoning Dataset and Hybrid Multimodal Fusion Model for Visual Question Answering in Vietnamese}                      



%
\author[1,2]{Khiem Vinh Tran}[
                        orcid=0000-0001-7511-2910]


\ead{khiemtv@uit.edu.vn}


\credit{Conceptualization, Methodology, Data curation, Software, Writing - original draft}

\affiliation[1]{organization={University of Information Technology},
    city={Ho Chi Minh city},
    country={Vietnam}}
\affiliation[2]{organization={Vietnam National University},
    city={Ho Chi Minh city},
    country={Vietnam}}
\affiliation[3]{organization={HUTECH University},
    city={Ho Chi Minh city},
    country={Vietnam}}

\author[3]{Hao Phu Phan}[
                        orcid = 0009-0000-3962-0117
                        ]
\ead{phuhao.p1004@gmail.com}
\credit{Methodology, Software}

\author[1,2]{Kiet Van Nguyen}[
                        orcid=0000-0002-8456-2742]
\ead{kietnv@uit.edu.vn}
\credit{Writing - review & editing}
\author[1,2]{Ngan Luu Thuy Nguyen}[
                        orcid=0000-0003-3931-849X]
                        
\credit{Supervision}
\cormark[1]

\ead{ngannlt@uit.edu.vn}

\credit{Data curation, Writing - Original draft preparation}




\cortext[cor1]{Corresponding author}



\input{tex/abstract}



\maketitle

\input{tex/1-Introduction}

\input{tex/2-Related}
\input{tex/3-Dataset}
\input{tex/4-Method}
\input{tex/5-Experiment}
\input{tex/6-Discussion}
\input{tex/7-Conclusion}

\appendix
\section{Several instances of ViCLEVR on question category.}
\begin{figure*}[!h]
    \centering
    \includegraphics[width=1\linewidth]{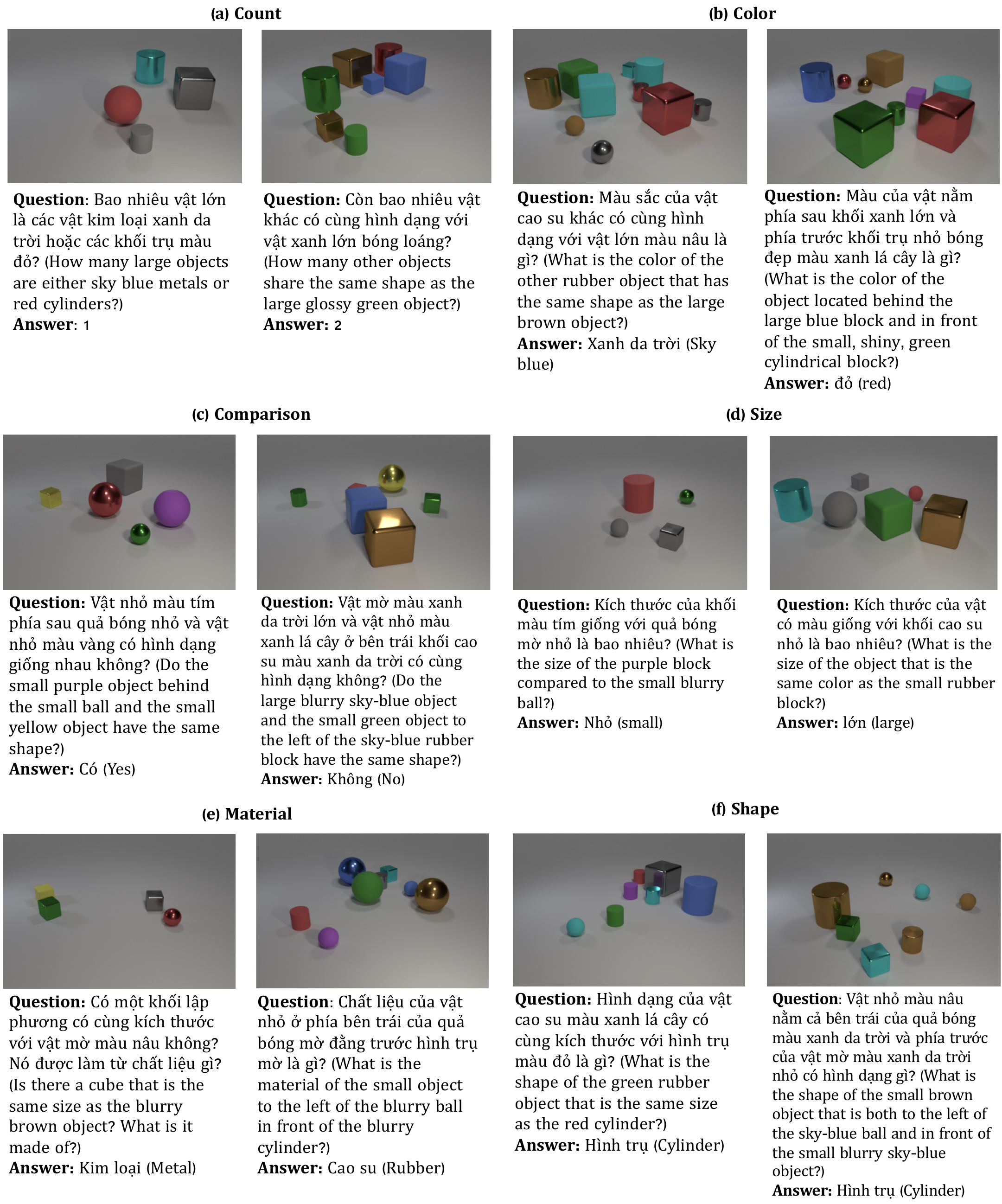}
    \caption{Examples of our dataset for question category.}
    \label{fig:AppendixCategory}
\end{figure*}

In \autoref{fig:AppendixCategory}, we present several exemplar instances from the ViCLEVR dataset, judiciously selected to illustrate the manner in which our queries harness question category factors pertinent to visible objects. Concurrently, we exhibit the corresponding responses, which encompass terminologies representative of such distinctive factors. These instances delineate the intricate interplay between categorical elements and their visible counterparts, serving to elucidate the multidimensional nature of the interrelations underpinning the queries and their respective resolutions within the framework of visual reasoning.
\section{Several instances of ViCLEVR on linguistic question type.}
\begin{figure*}[!h]
    \centering \includegraphics[width=1\linewidth]{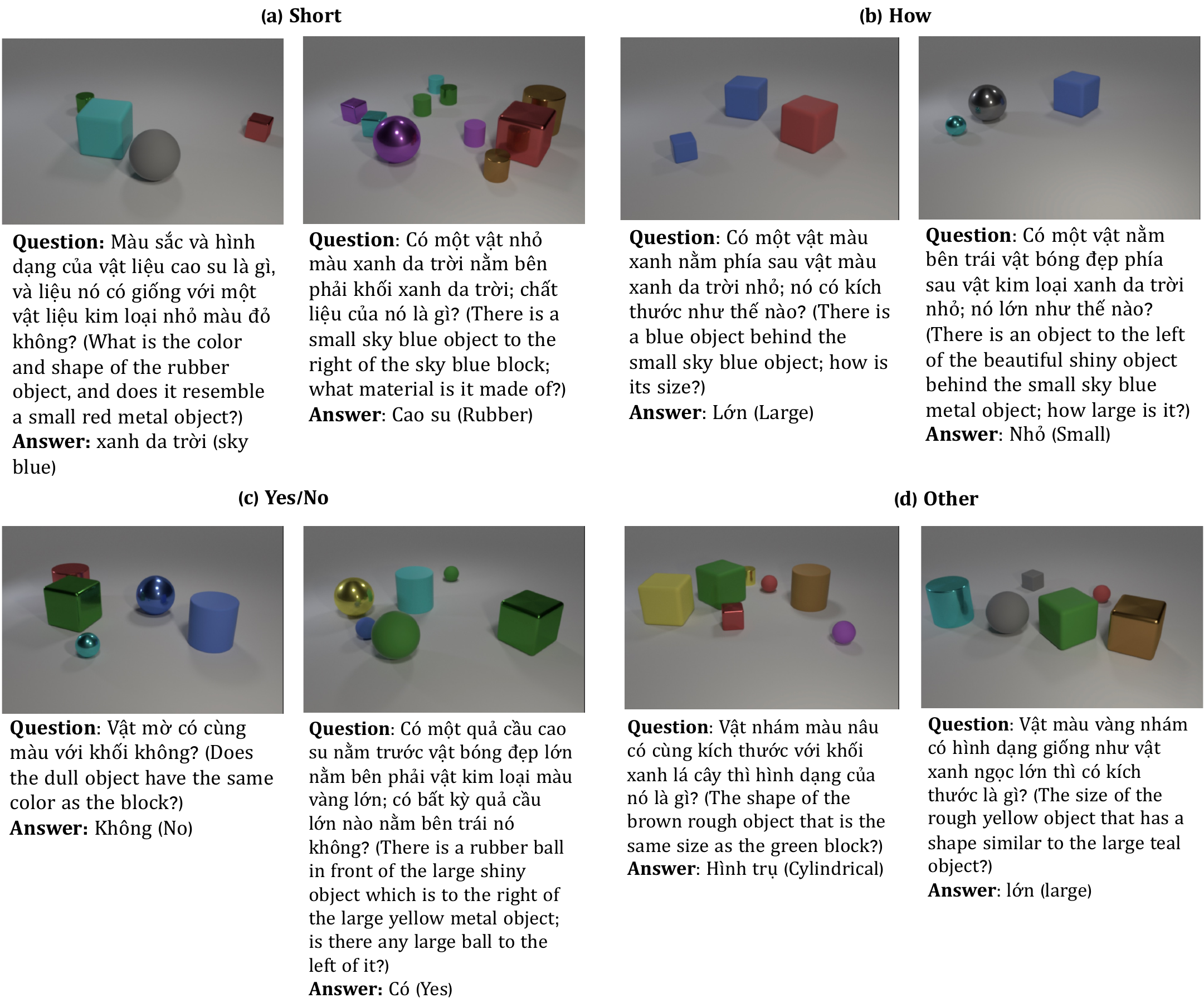}
    \caption{Examples of our dataset for linguistic question type.}
    \label{fig:AppendixType}
\end{figure*}

Within \autoref{fig:AppendixType}, multiple exemplary instances are displayed from the ViCLEVR dataset. These have been meticulously chosen to delineate the methodologies whereby our queries employ linguistic question types, associating them with factors intrinsic to visible entities. Simultaneously, the allied responses are showcased, incorporating terminologies that are indicative of such unique factors. 

These exemplifications illuminate the sophisticated synergy between categorical constructs and their perceptible equivalents, elucidating the multi-faceted relational dynamics inherent to the inquiries and their concomitant resolutions within the paradigm of visual reasoning. Such instances afford insight into the nuanced interdependencies that characterize the interface between linguistic articulation and visual perception, reflecting the complex, integrative nature of cognitive processing within this domain.
\bibliographystyle{cas-model2-names}

\bibliography{cas-refs}





\end{document}

%% file: tex/abstract.tex
\begin{abstract}
In recent years, Visual Question Answering (VQA) has gained significant attention for its diverse applications, including intelligent car assistance, aiding visually impaired individuals, and document image information retrieval using natural language queries. VQA requires effective integration of information from questions and images to generate accurate answers. Neural models for VQA have made remarkable progress on large-scale datasets, with a primary focus on resource-rich languages like English.
To address this, we introduce the ViCLEVR dataset, a pioneering collection for evaluating various visual reasoning capabilities in Vietnamese while mitigating biases. The dataset comprises over 26,000 images and 30,000 question-answer pairs (QAs), each question annotated to specify the type of reasoning involved. Leveraging this dataset, we conduct a comprehensive analysis of contemporary visual reasoning systems, offering valuable insights into their strengths and limitations.
Furthermore, we present PhoVIT, a comprehensive multimodal fusion that identifies objects in images based on questions. The architecture effectively employs transformers to enable simultaneous reasoning over textual and visual data, merging both modalities at an early model stage. The experimental findings demonstrate that our proposed model achieves state-of-the-art performance across four evaluation metrics. The accompanying code and dataset have been made publicly accessible at \url{https://github.com/kvt0012/ViCLEVR}. This provision seeks to stimulate advancements within the research community, fostering the development of more multimodal fusion algorithms,  specifically tailored to address the nuances of low-resource languages, exemplified by Vietnamese.
\end{abstract}

%% file: tex/1-Introduction.tex
\section{Introduction}
\label{sec:intro}
Visual question answering (VQA) has emerged as a highly challenging task within the field of Artificial Intelligence (AI), attracting significant attention from researchers. The objective of VQA is to predict an answer based on an input image and a corresponding question. It serves as a fundamental component for various complex AI applications, encompassing the automatic understanding of both offline and real-time video streams. Examples of such applications include assistive technologies for visually impaired individuals, collaborative robotics, and embodied intellectual assistants.

VQA presents a multimodal challenge that necessitates the seamless integration of fine-grained image analysis techniques with advanced natural language models. One of the primary hurdles in VQA \citep{IPM2022} lies in addressing the symbol grounding problem, which remains an unresolved issue in AI. In essence, this problem revolves around establishing meaningful connections between symbols within an AI model and real-life objects and situations. In the context of VQA \citep{CVIU20232}, the challenge involves mapping the symbols employed in natural language processing models, which interpret questions, to the objects and situations depicted in visual scenes processed by computer vision models.

The field of VQA predominantly focuses on a select number of high-resource languages, with English being the primary focus, which overlooks the diverse linguistic landscape represented by billions of speakers \citep{bender2019benderrule}. Data-intensive deep learning systems have led to significant advancements in VQA performance for high-resource languages. However, the lack of extensive datasets for low-resource languages presents a formidable challenge in their NLP processing \citep{ruder20194, hedderich-etal-2021-survey}. Consequently, addressing VQA in low-resource scenarios has become one of the foremost open challenges in the field of VQA today.

The challenges faced by languages with limited linguistic resources, such as Vietnamese, should indeed be recognized. Despite being the national language of Vietnam and spoken by nearly 100 million people, making it the 15th most widely spoken native language globally, Vietnamese still encounters resource scarcity, which poses obstacles in various artificial intelligence domains, including visual question answering research and development \footnote{https://www.worldometers.info/world-population/vietnam-population/}.

Research on Visual Question Answering (VQA) has predominantly focused on well-resourced languages, especially English, since its emergence around 2015 \citep{DBLP:conf/iccv/AntolALMBZP15}. Nevertheless, there exists a significant research lacuna concerning VQA in languages with limited resources, such as Vietnamese. Addressing this void, Tran et al. (2021) \citep{tran-etal-2021-vivqa} unveiled the ViVQA dataset, pioneering the development of a VQA dataset specifically for the Vietnamese language. However, it's imperative to acknowledge that the ViVQA dataset encapsulates only a fragment of the VQAv2 dataset since it was constructed utilizing a machine translation approach.

Taking into consideration the conceivable complications associated with validation and the intrinsic constraints of outcomes derived from machine translation, the efficacy of the semi-automated approach introduced in \citep{tran-etal-2021-vivqa} may not ascertain performance at a human level in the translation of an English VQA dataset into Vietnamese. As a result, the credibility of the ViVQA dataset as a benchmark may be compromised for executing experimental procedures, assessing VQA systems, or fostering advancements in Vietnamese VQA research.

Therefore, there is a clear and urgent need for a new semi-automatically annotated VQA dataset that can serve as a robust benchmark for VQA research in Vietnamese, addressing the limitations and potential issues associated with the existing ViVQA dataset. This effort would contribute significantly to the development and advancement of VQA in low-resource languages like Vietnamese.
Additionally, our research endeavors encompass the execution of a series of experiments, predicated on the methodologies and frameworks outlined in existing scholarly works. These experiments are conducted utilizing contemporary approaches, with a view to assessing and corroborating the validity and efficacy of current theories and models. Subsequent to this rigorous examination, we advocate for a novel hybrid multimodal fusion strategy, a conceptual framework conceived to synthesize existing approaches, with the aspiration of realizing superior empirical outcomes. 

The newly proposed paradigm serves as a substantial and robust baseline, instrumental for advancing subsequent inquiries and scholarly explorations, specifically within the context of our distinctive dataset and, more broadly, within the realms of Visual Question Answering (VQA) and Visual Reasoning tasks. The methodological innovation proffered by this research holds significant implications for the proliferation of knowledge and the inception of novel investigative trajectories within the interdisciplinary domain of visual cognition and computational reasoning, hence contributing to the cumulative progression of the academic discourse in this scientific field. Our contributions are as follows:
\begin{itemize}
    \item Firstly, we introduce the ViCLEVR dataset, which serves as a data source for Vietnamese Visual Question Answering (VQA). This benchmark encompasses four independent metrics, allowing for a comprehensive evaluation and providing insights into the performance of VQA systems. Additionally, the ViCLEVR benchmark can be utilized to assess part-based reasoning capabilities.
    \item Secondly, we introduce a novel visual reasoning methodology incorporating a hybrid multimodal fusion mechanism, which integrates elements of the Vietnamese language to bolster reasoning capabilities in tasks related to visual understanding. This innovative approach exploits the distinctive attributes inherent to the Vietnamese language, aiming to augment the precision and efficacy of visual reasoning processes. The integration of linguistic elements serves to enhance the nuanced understanding of visual stimuli, facilitating advanced interpretative and analytical insights, thereby contributing to the evolving discourse in the field of computational visual cognition and reasoning.
    \item Thirdly, we conduct a detailed analysis of five existing methods, alongside our novel approach, to investigate the impact of different model designs on the performance of VQA specifically for Vietnamese. Through this analysis, we gain valuable insights into the strengths and limitations of these methods, as well as our own proposed approaches, in the context of Vietnamese VQA.
\end{itemize}

This manuscript is structured in the subsequent manner. In \autoref{sec:related}, we delve into the existing methodologies prevalent in the field. In \autoref{sec:data}, the paper's principal contributions, specifically the development of a Vietnamese reasoning dataset and its analytical aspects, are accentuated. Next,  \autoref{sec:proposedmethod} delineates the theoretical underpinnings germane to the proposed methodology. A comparative assessment of the proposed solution vis-à-vis other extant baselines using the ViCLEVR dataset is furnished in \autoref{sec:ex}. Finally, the study concludes with reflective observations in \autoref{sec:conclude}.

\begin{figure*}
    \centering
    \includegraphics[width=1\linewidth]{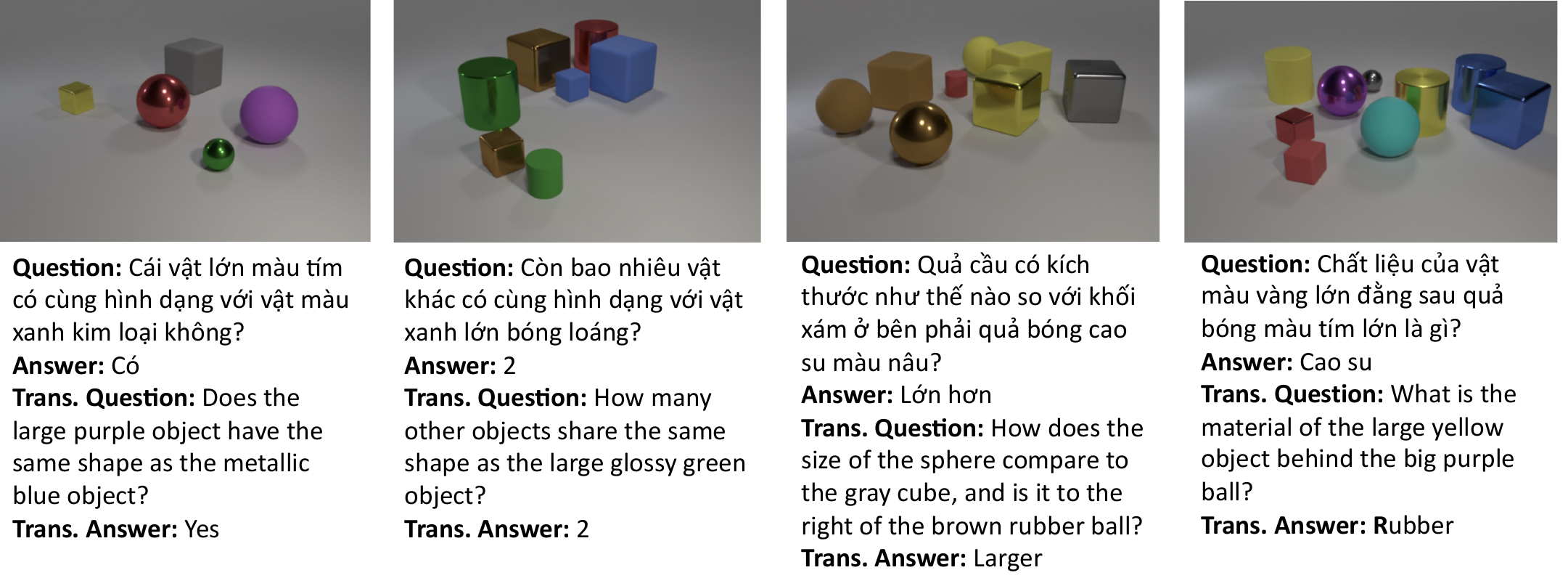}
    \caption{Typical examples of our proposed dataset}
    \label{fig:enter-label}
\end{figure*}

%% file: tex/2-Related.tex
\section{Related work}
\label{sec:related}
\begin{figure*}[!h]
    \centering
    \includegraphics[width=0.7\linewidth]{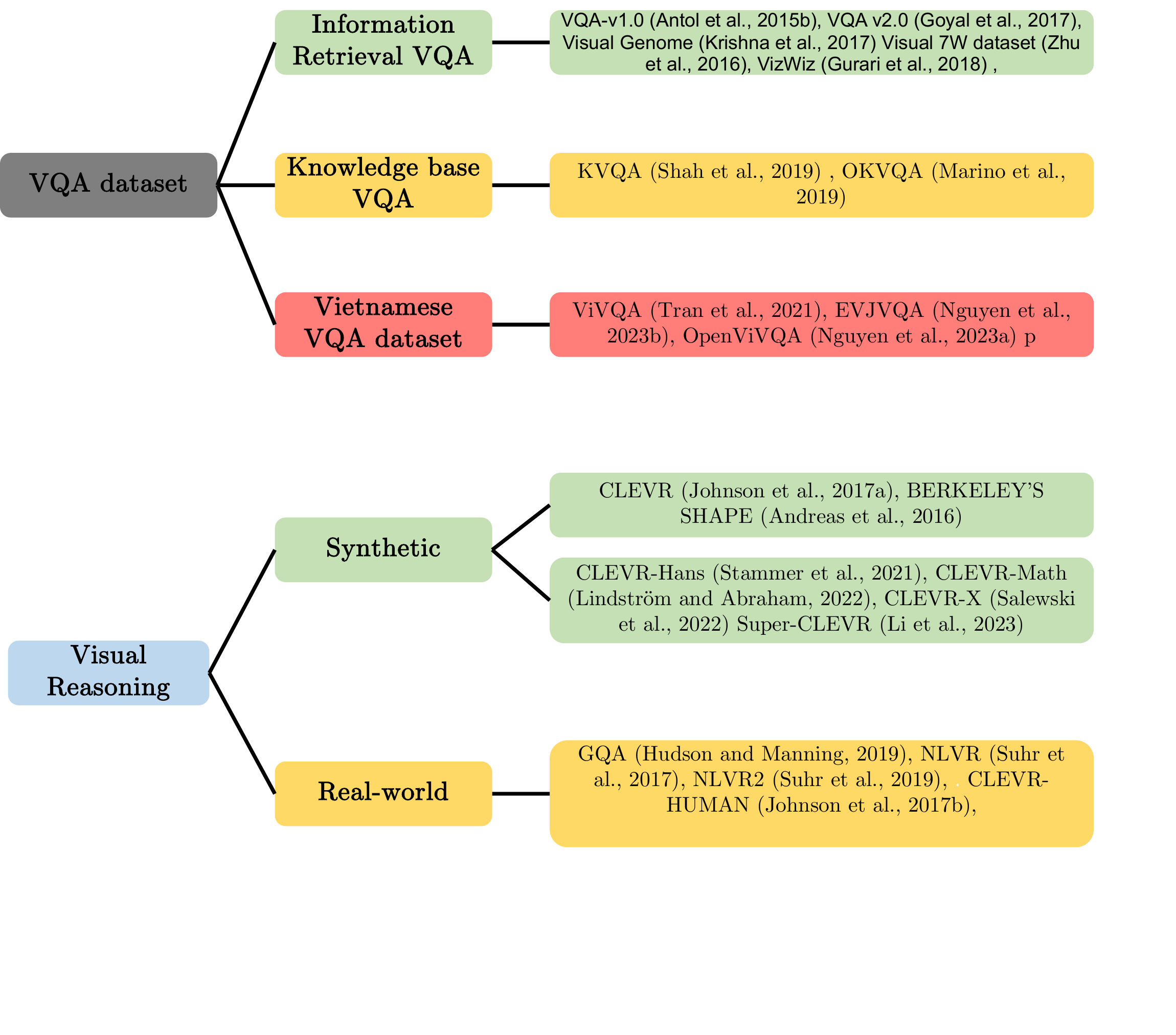}
    \caption{An illustration of dataset approach in VQA and Visual Reasoning}
    \label{fig:overview}
\end{figure*}

\sstitle{VQA datasets}
VQA-v1.0 \citep{antol2015vqa} is a well-established Visual Question Answering (VQA) dataset that utilizes the COCO dataset \citep{ren2015exploring}. It comprises two distinct subsets: VQA-v1-real, featuring real photographs, and VQA-v1.0-abstract, incorporating artificially generated cartoon images. The training phase of VQAv1-real utilizes 123,287 images, while the testing phase employs 81,434 images selected from the COCO dataset.

VQA v2.0 \citep{goyal2017making} represents an updated iteration of the VQA dataset, designed to address previous concerns and biases. The training set consists of 443,757 image pairs, the validation set includes 214,354 image pairs, and the test set encompasses 447,793 image pairs. Remarkably, this revised version is twice the size of its predecessor. The dataset comprises 1.1 million pairs of images and questions, accompanied by 13 million associated answers, all of which have been annotated arbitrarily.


Visual Genome \citep{gnome2017} is a dataset that aims to enhance cognitive abilities, specifically spatial reasoning, through the practice of genome visualization. The dataset comprises a large collection of over 108,000 photos, each featuring an average of 35 distinct items. These items are associated with 26 characteristics, and there exist 21 pairwise interactions between objects within the photo. The primary objective of Visual Genome is to facilitate improved performance in cognitive tasks involving spatial connection thinking.

The Visual 7W dataset \citep{zhu2016visual7w} is derived from the comprehensive Visual Genome dataset \citep{gnome2017} and focuses on a subset of the available data. It consists of 327,939 question-answer pairs and 47,300 images sourced from MS COCO \citep{DBLP:conf/eccv/LinMBHPRDZ14}. The dataset includes 1,311,756 multiple-choice questions, carefully categorized into seven distinct question types: what, where, when, who, why, how, and which, collectively forming the "7W" classification.

VizWiz \citep{gurari2018vizwiz} is a groundbreaking vision dataset, obtained from individuals who are blind, making it the first widely accessible dataset of its kind. It presents an intriguing challenge within the field of Visual Question Answering (VQA) \citep{CVIU2019} by focusing on the prediction of whether a given visual question can be answered. The dataset builds upon previous research \citep{47094} and encompasses a collection of 72,205 visual questions, accumulated over a span of four years. These questions were collected through the VizWiz mobile application, available on both iPhone and Android platforms.

The KVQA (Knowledge-aware VQA) dataset \citep{Shah_Mishra_Yadati_Talukdar_2019} was curated with a specific focus on questions that necessitate external knowledge for accurate answers. It contains a total of 183,000 question-answer pairs, with approximately 18,000 individuals captured within 24,000 images. Answering the questions in this dataset requires employing multi-entity, multi-relation, and multi-hop reasoning over a Knowledge Graph (KG) \citep{IPM2022}. Another distinctive aspect of this dataset is the presence of inquiries that extend beyond KG entities as ground-truth answers.

OKVQA dataset \citep{DBLP:conf/cvpr/MarinoRFM19}  is the most extensive knowledge-based Visual Question Answering (VQA) dataset available, featuring comprehensive annotations, including questions, answers, and knowledge categories. This dataset comprises 14,031 images accompanied by 14,055 diverse questions covering a wide array of topics, such as travel, materials, sports, cooking, geography, plants, animals, science, weather, and many others.

\sstitle{Visual reasoning in VQA} The CLEVR dataset \citep{Johnson_2017_CVPR} is designed as a diagnostic tool for the evaluation of explicit visual reasoning abilities \citep{ZHENG202114} within the context of Visual Question Answering (VQA). The dataset comprises an extensive collection of 100,000 images, accompanied by 864,968 associated questions. Ground-truth annotations are provided for the photographs, encompassing essential item properties such as size, shape, material, color, and spatial coordinates.

Berkeley's SHAPES dataset \citep{andreas2016neural} presents a valuable resource for the investigation of explicit reasoning in visual question-answering tasks. It consists of synthetic images featuring 2D abstract shapes. The dataset includes 15,616 synthetic pictures, each encompassing diverse sizes and spatial placements, alongside a set of 244 binary questions requiring responses in a yes or no format.

The GQA dataset, developed by Stanford \citep{hudson2019gqa}, is a recent addition to their collection, specifically designed to facilitate scene comprehension and reasoning tasks. It comprises a comprehensive set of 113,018 real-world images sourced from the Visual Genome dataset, accompanied by their corresponding scene graphs. The scene graphs are subjected to extensive normalization and rectification processes to ensure precise annotations and generate high-quality queries.
In addition to the images and scene graphs, the GQA dataset includes a vast collection of 22,669,678 multi-step questions. These questions are generated using a sophisticated question engine that leverages the rich information extracted from the scene graphs. Moreover, the dataset encompasses 524 structural patterns, with 250 patterns manually constructed and an additional 274 patterns retrieved from the VQA 1.0 dataset. These structural patterns serve as valuable resources for guiding the generation of diverse and challenging questions within the GQA dataset.

The NLVR dataset \citep{DBLP:conf/acl/SuhrLYA17} is a multimodal dataset that combines human language descriptions with synthetic visuals. The images feature various objects, such as triangles, circles, and squares, arranged in different sizes and positions within the image. The dataset includes hand-written descriptions for each image, provided by crowd workers. The NLVR2 dataset \citep{suhr-etal-2019-corpus} was designed to overcome language bias and improve upon the limitations of the original NLVR dataset, which was synthetic in nature. The NLVR2 dataset, also known as Natural Language for Visual Reasoning, includes pairs of visuals along with corresponding grounded natural language descriptions, similar to NLVR. By incorporating real-world images and more diverse language, NLVR2 aims to address issues such as restricted expressivity and semantic diversity encountered in the synthetic NLVR dataset.


In accordance with CLEVR, a proliferation of variant datasets has been generated, eliciting heightened interest and participation from scholars in the field. CLEVR-HUMAN \citep{DBLP:conf/iccv/JohnsonHMHFZG17} is tailored to the collection of human-generated free-form natural language queries concerning CLEVR images. CLEVR-Hans \citep{DBLP:conf/cvpr/StammerSK21} represents an innovative visual scene dataset characterized by its intricate portrayal of complex compositions involving diverse objects. This dataset further categorizes CLEVR images into multiple discrete classes, facilitating granular investigations.
CLEVR-Math \citep{DBLP:conf/nesy/LindstromA22} introduces a multimodal math word problems dataset, encompassing straightforward mathematical word problems primarily involving addition and subtraction. These problems are elucidated through a hybrid representation comprising textual descriptions and complementary images, effectively illustrating the contextual scenario. CLEVR-X \citep{Salewski2022} extends the foundational CLEVR dataset with the incorporation of natural language explanations \citep{ARRAS202214}, enhancing the dataset's interpretability and overall utility.
Super-CLEVR \citep{DBLP:conf/cvpr/LiWSKMDY23} emerges as a comprehensive initiative, systematically addressing diverse facets within the domain of Visual Question Answering (VQA) \citep{CVIU2023}. This dataset introduces four pivotal factors for examination, encompassing visual complexity, question redundancy, concept distribution, and concept compositionality, with the aim of advancing the collective understanding of these critical dimensions.

Nonetheless, despite the availability of these invaluable resources, a noticeable gap is discernible within the realm of visual reasoning datasets akin to CLEVR, particularly within the context of widely spoken yet low-resource languages, exemplified by the Vietnamese language. This conspicuous lacuna within the research landscape serves as a compelling catalyst propelling our ongoing efforts to construct a novel CLEVR-style dataset, meticulously tailored to prevalent low-resource languages, with Vietnamese serving as a prominent exemplar, thus redressing this unmet need.

\sstitle{Visual Question Answering Datasets in Vietnamese}
While there are myriad benchmarks for the Visual Question Answering (VQA) task in English \citep{CVIU2017}, languages with scarce linguistic resources, such as Vietnamese, face a notable dearth of such resources. In a significant stride in 2021, Tran et al. \citep{tran-etal-2021-vivqa} launched the ViVQA dataset, marking the inception of a dedicated VQA dataset for Vietnamese. To craft this dataset, machine translation techniques were harnessed to transcribe questions and answers from a segment of the VQAv2 dataset into Vietnamese, followed by an exhaustive verification to ensure the accuracy and fluency of the translations.
 
Building on these foundational steps, Nguyen et al. in 2022 \citep{nguyen2023vlsp} broke new ground by unveiling a multilingual dataset through a shared task. Notably, this dataset incorporates the Vietnamese language, broadening the scope of VQA research to delve into the Vietnamese linguistic setting. This seminal work signifies a monumental leap in VQA research, especially tailored to the unique linguistic nuances and demands of Vietnamese.

Moreover, in a subsequent study, Nguyen et al. \citep{NGUYEN2023101868} presented the OpenViVQA (Open-domain Vietnamese Visual Question Answering) dataset. This extensive compilation is curated for VQA tasks that necessitate open-ended responses in Vietnamese and comprises over 11,000 images paired with more than 37,000 question-answer sets.

To the best of our knowledge, there remains an unfulfilled need for a dataset in Vietnamese that zeroes in on visual reasoning. Motivated by this void, our research aims to bridge this lacuna and augment the domain of visual reasoning tailored to the Vietnamese linguistic context.

%% file: tex/3-Dataset.tex
\section{ViCLEVR  dataset}
\label{sec:data}
ViCLEVR provides a dataset posing challenges that necessitate advanced reasoning skills for effective resolution. It acts as a pivotal tool for performing extensive diagnostic studies, focusing on discerning the depth of visual reasoning proficiencies inherent in Visual Question Answering (VQA) systems. For meticulous management and integrity of the dataset, it employs synthetic images and auto-generated questions.

Every image in the dataset is paired with accurate object locations and attributes, offering exact and dependable referential data. Questions contained within the dataset are also rendered in a format that is machine-readable, enabling methodical analysis and assessment. The presence of these ground-truth configurations permits diverse analytical approaches, including evaluations based on the type of question, the topology of the question (examining chain versus tree structures), the length of the question, and varied object relationships. Such in-depth examinations aid in acquiring a holistic comprehension of the competencies and performance levels of VQA models.
\subsection{Overview}
Our dataset consists of 26,000 rendered images from CLEVR dataset \citep{Johnson_2017_CVPR} and 30,000 semi-auto-annotated questions. The images are rendered from a synthetic scene with fixed objects and materials. The questions are generated using a grammar that allows for a wide range of compositional queries. Motivated by CLEVR \citep{Johnson_2017_CVPR}, the questions in our dataset are divided into six categories:
\begin{figure*}[!h]
    \centering
    \includegraphics[scale=0.4]{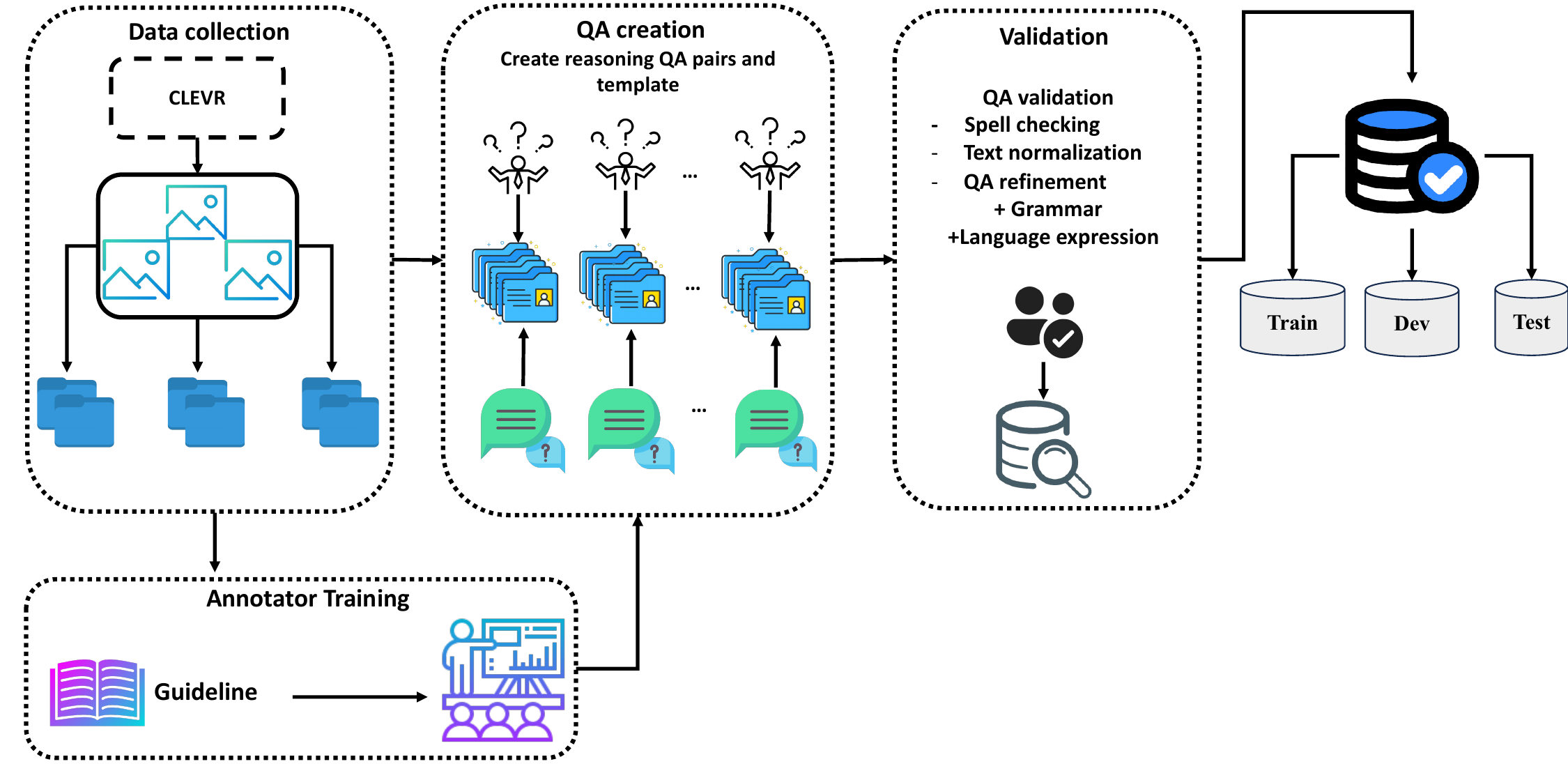}
    \caption{The overview process of creating our
dataset ViCLEVR }
    \label{fig:quescreate}
\end{figure*}

\begin{itemize}
    \item Counting: These questions ask how many objects of a particular type exist in the image.
    \item Color: These questions ask which color of the particular object exists in the image.
    \item Comparison: These questions ask to compare two objects in the image.
    \item Size: These questions ask for the size of an object in the image.
    \item Material: These questions ask about the material of an object in the image.
    \item Shape: These questions ask about the shape of an object in the image.
\end{itemize}
To enhance the evaluative process, we provide meticulous annotations for each question, offering explicit descriptions of the reasoning processes and approaches necessary to arrive at the correct answer. These annotations serve as an integral component in evaluating the performance of visual question answering (VQA) models.

\subsection{Question-Answer Pair Creation}
The CLEVR dataset formulates questions using a template-driven methodology, leveraging what are termed as "question families" to automate question generation. Within CLEVR, there exist 90 distinct question families. Each of these families is characterized by a singular program template and, on average, contains four text templates. The derivation of these text templates is twofold: initial templates are manually crafted with one or two templates allocated per family, and the remainder are sourced through crowdsourced rephrasing of questions. To amplify linguistic variability, alternative terms describing shape, color, and material are integrated. Remarkably, by employing templates that house up to 19 variables, these limited question families are capacitated to produce an extensive array of distinct questions.

However, it should be noted that the aforementioned approach is not directly applicable to generating Vietnamese questions due to the grammatical differences in the structure of the Vietnamese language. Consequently, an alternative approach is required for creating Vietnamese questions within the CLEVR framework, drawing inspiration from the construction process employed in the CLEVR \citep{Johnson_2017_CVPR} and GQA datasets \citep{hudson2019gqa}.

In order to complete this phase, we enlist a team of proficient Vietnamese crowd workers. By harnessing the advantages of crowdsourcing, our goal is to amass a substantial volume of data with the desired variations that ensure linguistic diversity in the generation of authentic questions and answers, along with a comprehensive vocabulary. The dataset's quality is of utmost significance, and to uphold it, we have established meticulous guidelines for monitoring and maintaining the dataset's integrity.

 Crowd workers are furnished with a series of protocols, delineated in Table \ref{tab:annotatedProcess}, to which compliance is mandatory. To preserve the exploratory essence of the questions and answers, it is anticipated that the questions would predominantly aim at eliciting information rather than manifesting a binary or selective characteristic. In a parallel manner, answers are envisaged to be extensive, surpassing the scope of single-word replies. Additionally, meticulous attention is applied to the regulation of queries concentrating on quantities, colors, and orientations. Although pivotal in discerning and differentiating objects, these elements may be prone to linguistic divergences and inconsistencies during the crowdsourcing phase. Illustrations embodying these norms are depicted in Figure \ref{fig:quescreate}.

Crowd workers are allocated the responsibility of formulating question-answer (QA) pairs corresponding to an optimum array of images. In scenarios where the cumulative number of QAs associated with an image is below the stipulated minimum threshold articulated in the guidelines, crowd workers are accorded the latitude to generate QAs reflecting analogous semantics to the extant ones, predominantly in the context of images exhibiting restrained intricacies. Conversely, in instances where an image is bereft of specificities and recognizable nuances, it may be excluded from the procedure. The generation of questions and answers persists until the fulfillment of all predetermined subsets.
\begin{table*}[!h]
    \caption{The formulation of questions and answers stemming from the ViCLEVR dataset complies with a designated set of protocols, intended to maintain uniformity and precision across the dataset. This adherence to established guidelines assures the reliability and consistency of the data, thereby contributing to the credibility of research outcomes derived from it.}
    \centering

    \begin{tabular}{c|c}
    \hline
        \textbf{No.} & \textbf{Description}    \\ \hline
        1 &Each image in the dataset should have a minimum of 1 associated question. \\ 
        2  & Each question should have a unique and corresponding answer. \\ 
        3  & The questions and answers should focus solely on the activities and objects depicted in the image. \\ 
        4  & Annotators are required to refrain from including personal opinions or emotions while annotating.  \\ 
         5 & The questions should encompass diverse reasoning approaches and objectives from different perspectives.  \\ 
         6 & The question types should not be limited to yes/no or selective questions. \\

    \hline
    \end{tabular}
\label{tab:annotatedProcess}
\end{table*}
\subsection{Dataset Validation}

In order to maintain a high quality and consistent dataset, we subject it to a rigorous validation process, which is depicted as one of the steps in the pipeline illustrated in \autoref{fig:quescreate}. Initially, a skilled crowd worker is assigned a portion of subsets from the dataset and tasked with identifying and rectifying any spelling or syntax errors they encounter. This process aims to enhance the overall accuracy and linguistic integrity of the dataset.

For optimizing the subsequent training phase, the dataset's question-answer (QA) pairs undergo preprocessing. This entails transforming the text into lowercase and introducing suitable whitespaces between words and punctuation. Such preprocessing measures foster uniformity and consistency in the textual information, guaranteeing its alignment with the training models and algorithms.
\subsection{Data Analysis}
The ViCLEVR  dataset encompasses a total of 26,000 images, each accompanied by 30,000 question-answer pairs that pertain to the visual content of the images. To ensure unbiased evaluation, we partition the dataset into training, test, and validation sets with a randomized distribution, adhering to a ratio of 7:2:1, respectively.
\begin{figure}
    \centering
    \includegraphics{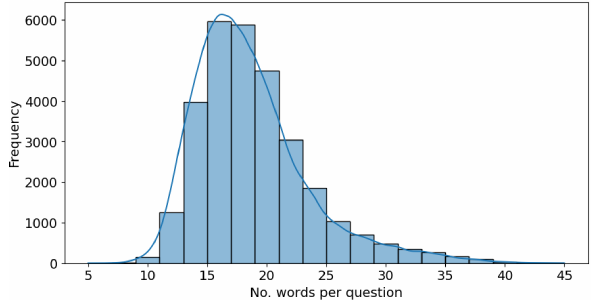}
    \caption{The distribution of question length in the ViCLEVR dataset.}
    \label{fig:queslen}
\end{figure}
\begin{figure}
    \centering
    \includegraphics{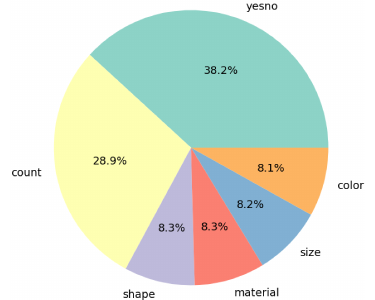}
    \caption{The distribution of question types in the ViCLEVR dataset.}
    \label{fig:questype}
\end{figure}

\begin{table*}[!h]
    \caption{The statistical analysis of ViCLEVR reveals that a significant proportion of the questions exhibit distinct characteristics, with only a limited number of questions from the validation and test sets being included in the training set.}
    \centering

    \begin{tabular}{c|c|c|c|c}
    \hline
        \textbf{Split} & \textbf{Images}  & \textbf{Questions} &\textbf{Unique Questions} &\textbf{Overlap with train}  \\ \hline
        Train &18,430&21,000& 20,323 & -  \\ 
        Dev  &5044 &6000& 5892 & 134 \\ 
        Test  & 2742&5000& 2964 &96\\ 
        Total  & 26,216& 30,000& 29,179&-\\

    \hline
    \end{tabular}
\label{tab:analysis}
\end{table*}

\begin{table*}[!h]
\caption{The comparison between ViCLEVR and existing Vietnamese VQA datasets}
\begin{tabular}{cccccc}
    \hline
        \textbf{Dataset} & \textbf{Task}  & \textbf{Question} &\textbf{Image} & \textbf{Source of Question} & \textbf{Source of Image}  \\ \hline
        ViVQA \citep{tran-etal-2021-vivqa} &VQA&   15,000 & 10,328  & Automation & VQAv2 \\
        OpenViVQA \citep{NGUYEN2023101868} &VQA & 37,000 & 11,000 & Crowdsourcing & Crowdsourcing \\
        ViCLEVR (Ours) &Visual Reasoning& 30,000 & 26,216 & Semi-automation & CLEVR \\ 
       
    \hline
    \end{tabular}
\end{table*}

\begin{table*}[!h]
    \caption{A comparative analysis of linguistic levels is conducted among Vietnamese VQA dataset. }
    \centering

    \begin{tabular}{lllllllllll}
\hline
                & \multicolumn{3}{c}{\textbf{Word}}              & \multicolumn{3}{c}{\textbf{Dependency}}           & \multicolumn{3}{c}{\textbf{Height}}             \\
\textbf{Dataset} & \textbf{min} & \textbf{mean} & \textbf{max} & \textbf{min} & \textbf{mean} & \textbf{max} &  \textbf{min} & \textbf{mean} & \textbf{max} \\ \hline
ViVQA \citep{tran-etal-2021-vivqa}        &  3      & 9.5        & 24       & 2    & 7.3       & 23        & 2   &      5.5 & 14            \\
OpenViVQA \citep{NGUYEN2023101868}   &    3       &     10.1        &       32     &      2     &   7.8         &         27    &    2     & 5.2 &  16                    \\
ViCLEVR (Ours)        &   5      &  18.57         &    45        &   5       &     16.77      &     40      &  2 & 3.88 & 10                                           \\ \hline
\end{tabular}

\label{tab:Linguisticannotated}
\vspace{0.5cm}
\begin{subtable}[t]{0.5\textwidth}
\centering 
\begin{tabular}{c|c|c|c}
    \hline
        \textbf{Dataset} & \textbf{Word}  & \textbf{Phrase} &\textbf{Sentence}  \\ \hline
        ViVQA \citep{tran-etal-2021-vivqa} &3,276&  6,321 &  0 \\
        OpenViVQA \citep{NGUYEN2023101868} &1,067 & 21,022 & 12,289 \\
        ViCLEVR (Ours) &557,131& 503,175 & 30,000  \\ 
       
    \hline
    \end{tabular}
\end{subtable}
       
\end{table*}

The distribution of question lengths within the ViCLEVR dataset is depicted in \autoref{fig:queslen}, showcasing the dataset's diversity and the intricate nature of its questions. Notably, a considerable proportion of the questions falls within the length range of 15 to 25. 
A statistical analysis of ViCLEVR, as shown in \autoref{tab:analysis}, reveals that a substantial portion of the questions exhibit distinct characteristics, while only a limited number of questions from the validation and test sets are present in the training set.

Within the scope of ViCLEVR, the linguistic attributes of the dataset entail a multitude of statistical dimensions, such as the frequency of questions, answers, and the semantic dependencies manifested within both questions and answers. The height of the semantic tree, structured based on these semantic dependencies, is also included, as detailed in Table \ref{tab:Linguisticannotated}.

The Linguistic Complexity Specification (LCS) methodology, as presented by \citep{NGUYEN2023101868}, serves to evaluate the intricacy of linguistic constructs within sentences. It probes into the statistical interplays among tokens present in specified sentences, taking cues from the outcomes of the corresponding dependency parser pertinent to the language in question. Leveraging the insights from dependency analysis, the LCS frames semantic structures, subsequently gauging their depths. Heightened counts of dependencies coupled with more extended semantic structures suggest augmented sentence complexity.

Concurrently, the Linguistic Level Specification (LLS) method \citep{NGUYEN2023101868} is harnessed to classify the introduced texts into categories such as individual words, compound phrases, or full sentences, contingent on the operational dependency parser. The foundational concept guiding this method posits that texts constituted by a lone token (segmented by word for Vietnamese or demarcated by spaces for English) are delineated as words. Those that feature a central token functioning as an action word, complemented by another token serving its subject, are demarcated as sentences. Residual texts are categorized as phrases. Implementing the LLS method facilitates discernment of the prevalent linguistic tier of sentences that humans conventionally opt for in response to queries. This underscores the inherent spontaneity and breadth in the dataset's answers, emphasizing the organic spectrum of human responses.

%% file: tex/4-Method.tex
\section{Our Proposed Model}
\label{sec:proposedmethod}
\begin{figure*}[!h]
    \centering
    \includegraphics[width=1\linewidth]{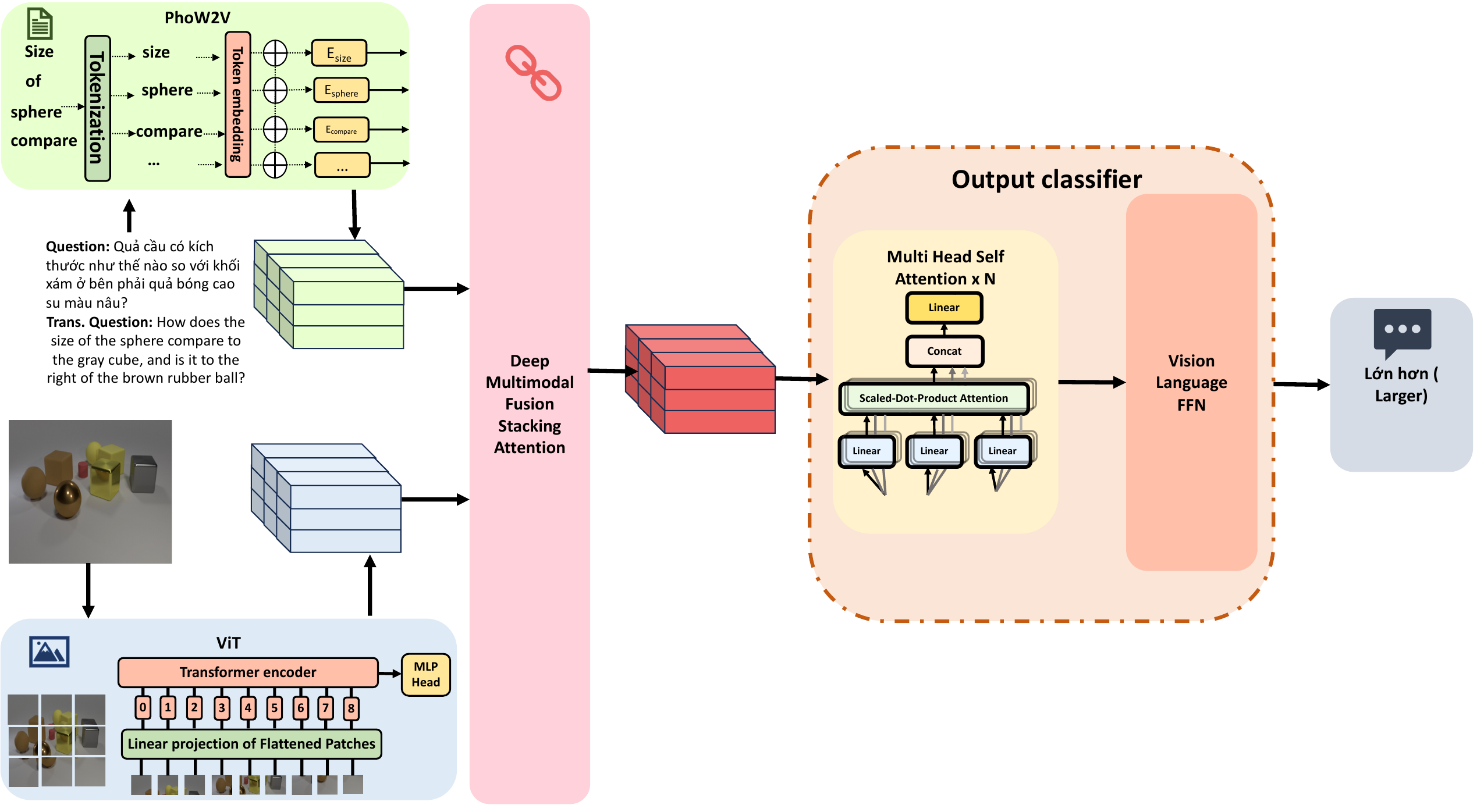}
    \caption{The overview architecture of PhoViT.}
    \label{fig:proposedModel1}
\end{figure*}

The method we propose is architecturally segmented into four principal segments: the Image Embedding module for assimilating visual information, the Question Embedding module for textual integration, the Multimodal Fusion module for amalgamating the extracted features, and the Classifier layer. The latter is integral as it is instrumental in forecasting the corresponding answers, a schematic of which is illustrated in \autoref{fig:proposedModel1}. The defining structural attributes of this model unfold as follows:
\subsection{PhoViT}
\subsubsection{Question embedding}
For each individual instance within the context of both training and testing, the input comprises a textual question and an associated image. The question undergoes a process of tokenization, wherein it is initially segmented into constituent words through the utilization of spaces and punctuation marks. Furthermore, numerical values or words grounded in numerical representations are encompassed within the scope of words within this context. Every individual word undergoes the transformation into a vector representation through the utilization of a look-up table. The entries within this table comprise 300-dimensional vectors, which are concurrently learned alongside other training parameters. It is worth noting that these vectors are initially initialized with pre-trained values.
In our study, we employ PhoW2V embeddings \citep{tuan-nguyen-etal-2020-pilot}, which are characterized by 300-dimensional word representations. These embeddings are generated using the Word2Vec skip-gram model \citep{DBLP:conf/nips/MikolovSCCD13}, specifically adapted for the Vietnamese language. Notably, these embeddings are derived from an extensive 20GB corpus of Vietnamese text.

The initial step involves tokenizing the input question into individual words, which is subsequently limited to a maximum length of 44 words. Each word within the question is then subject to conversion into a vector utilizing the 300-dimensional PhoW2V word embeddings, a resource that has been pre-trained on an extensive corpus. This process yields a sequence of words, constituting a matrix of dimensions $n\times 300$, where $n \in [1, 44]$ signifies the number of words present in the question. Importantly, the resulting question features $Q$ for all words are retained and collectively form a question feature matrix denoted as $Y \in R^{n\times d}$.
\subsubsection{Image embedding}
The input image undergoes processing via a standard Transformer architecture, drawing inspiration from the Vision Transformer (ViT) as described by \citep{dosovitskiy2021an}. The image, represented as \(e_x \in \mathbb{R}^{H \times W \times C}\), where \(H\) and \(W\) delineate its resolution and \(C\) indicates channel count, is restructured into a sequence of linearized 2D patches, expressed as \(x_p \in \mathbb{R}^{N \times (P^2 \cdot C)}\). The patch dimensions are denoted by \((P, P)\) and the total number of these patches is computed as \(N = \frac{HW}{P^2}\), which also defines the Transformer's input sequence length. Throughout the Transformer's layers, a consistent latent vector size \(D\) is retained. These patches are linearly projected to generate \(D\)-dimensional vectors (see Eq. 1), which results in the patch embeddings and the initial learnable embedding in the sequence. The final state of this embedding post-Transformer encoding becomes the image representation \(y\).

For both pre-training and subsequent fine-tuning, a classification head is integrated with the Transformer encoder. In the pre-training phase, this head is formulated as a Multi-Layer Perceptron (MLP) with one hidden layer, while during fine-tuning, a singular linear layer is employed. To maintain positional context, position embeddings are added to the patch embeddings. Our model opts for conventional 1D position embeddings due to negligible performance increments observed with intricate 2D-aware embeddings. The processed sequence of embedding vectors then serves as the input for the Transformer encoder, which aligns with the structure detailed by \citep{NIPS2017_3f5ee243}. This encoder is characterized by successive layers of multi-headed self-attention and MLP modules, with the ensuing output being mapped onto image features \(I\).
\subsubsection{Multimodal fusion}
The Multimodal Fusion module encompasses a Deep Stacked Attention module, akin to the approach presented in \citep{8953581}. Utilizing the previously mentioned question features denoted as Q and question features represented as I, we engage in profound co-attention learning by channeling the input features through a deep co-attention model comprising K deep attention layers (DA), arranged in a cascaded manner (denoted as $DA_1, DA_2, ..., DA_K$). Denoting the input features for the K-$th$ DA layer as $Q_{K-1}$ and $I_{K-1}$, their resultant output features are designated as $Q_K$ and $I_K$, respectively. These subsequently become the inputs for the next DA layer, following a recursive progression.
\begin{equation}
    [Q_K,I_K] = DA_K([Q_{K-1},I_{K-1}])
\end{equation}
The deep stacked fusion model incorporates a seamless alignment of \( K \) layers of Dual Attention (DA) \citep{CVIU20192} in a profound manner. This leads to the derivation of \( Q_K \) and \( I_K \) as the final attended features pertaining to image and question at the last layer. The resulting fused embedding delineates the representations of both the image and the question. Subsequently, this embedded fusion is streamlined and channeled towards the designated output classifier.
\subsubsection{Output classfier}
Subsequent to the deep stacking phase inherent to multimodal fusion, the ensuing fused features intrinsically embody expansive information pertaining to the attention distribution across the words of the question, denoted as $Q_w$, and the regions of the image, represented as $I_w$. In light of this, a model delineated for attentional reduction is constituted, integrating a multi-layer Multihead Self Attention (MSA) mechanism spanning N layers to procure its attended features $Q_f$ and $I_f$. To illustrate, considering $Q_w$, the attended feature $Q_f$ is derived as:
\begin{equation}
\alpha = softmax(MSA(Q_w))
\end{equation}
\begin{equation}
Q_f = \sum_{i=1}^{N} \alpha_i \times Q_{w_i}
\end{equation}
Herein, $\alpha = [ \alpha_1, \alpha_2, ..., \alpha_N ] \in  R^N$ represent the attention weights that are discerned through learning. This is analogous for $I_w$ and $Q_w$.
Utilizing the computed $Q_f$ and $I_f$, a linear multimodal fusion function is construed as:
\begin{equation}
    z = LayerNorm \left(W^T_x \times Q_f  + W^T_y \times I_f\right)
\end{equation}
Here, $W_x, W_y \in R^{d×df}$ serve as two linear projection matrices, with $df$ denoting the unified dimensionality of the fused feature. The fused feature $z$ is subsequently projected into a vector $s \in R^N$, where $N$ represents the count of the most prevalent answers within the training set, and is followed by the application of a sigmoid function.

In the following steps, the final pooled outputs corresponding to the two pairs in question and images are consolidated to forge a concatenated representation. This amalgamated representation is then channeled into a classifier layer, executing operations via the Vision-Language Feed-forward Network (VL-FFN) as a fully connected layer, and then used to forecast the associated answer.  In alignment with the findings of \citep{wang2023image}, leveraging an assortment of modality specialists augments the model's ability to assimilate a more extensive array of information specific to each modality. The consolidated self-attention module excels in discerning the alignments between various modalities, enabling enhanced integration for tasks characterized by their multimodal attributes, such as those involving vision and language. This methodology assists in the meticulous amalgamation of the nuanced elements inherent to each modality, leading to a more fortified and insightful harmonization and amalgamation of information across modalities. 

For simplicity, PhoViT is trained with binary cross-entropy (BCE) loss \citep{DBLP:conf/cvpr/TeneyAHH18}. Binary Cross-Entropy (BCE) is employed as the loss function to facilitate the training of an N-way classifier, which is constructed atop the fused feature \( z \). Throughout the inference phase, caption tokens are generated sequentially in an autoregressive fashion, enabling a coherent and contextually aware generation of caption components. This method ensures the holistic incorporation of information in the generation process, providing a sequentially refined output.

\begin{equation}
    L = -\sum_{i}^{N} \sum_{j}^{M} z_{ij} log(\widehat{z}_{ij}) - (1 - z_{ij}) \times log(1 - \widehat{z}_{ij}))
\end{equation}
In this context, the indices \(i\) and \(j\) respectively traverse through the \(M\) training questions and \(N\) candidate answers, systematically iterating each element within the specified sets to perform subsequent operations or comparisons, ensuring comprehensive evaluation or processing over the entire spectrum of training questions and candidate answers.

%% file: tex/5-Experiment.tex
\section{Experiments}
\label{sec:ex}
\subsection{Comparative baselines}
Conducting experiments encompassing the entirety of available methodologies presents logistical complexities. Consequently, we have undertaken the replication of a judiciously selected subset of methods. This subset includes baseline models that exclusively rely on textual information (LSTM-Q \citep{DBLP:conf/iccv/AntolALMBZP15}) . Additionally, we have implemented a straightforward baseline model that combines LSTM combined with CNN representations, a configuration known to approximate the state-of-the-art performance (LSTM+CNN \citep{Johnson_2017_CVPR}). Our subset also encompasses both historical and contemporary state-of-the-art approaches in ViVQA dataset (VieHieCoAtt \citep{tran-etal-2021-vivqa} and BARTPhoBEiT \citep{tran2023bartphobeit}). Finally, we have incorporated novel methodologies specifically developed for the Vietnamese language utilizing a recently introduced dataset (ViMCAN \citep{NGUYEN2023101868} ). 

Within the framework of the current experiment, the methodologies that have been employed are categorized into four unique types, to wit: Traditional Neural Network, Co-attention, Transformer, and Hybrid. Each typology is characterized by distinctive techniques and methodologies, specifically devised for addressing tasks related to Visual Question Answering (VQA).

The Traditional Neural Network typology is embedded with conventional deep learning methodologies, explicitly designed for the complexities inherent in VQA tasks. It serves as a foundational approach, focusing on traditional computational models to interpret and process visual and contextual information.

The Co-attention typology, on the other hand, is structured around the implementation of a co-attention mechanism, establishing a framework that allows parallel focusing and alignment on various segments of the input data, creating a cohesive interplay between the visual and linguistic components.

In the Transformer typology, the application is centered around the transformer model, a well-established approach known for its self-attention mechanism, offering a refined method for processing sequential data and enabling the model to prioritize different segments of input information based on contextual relevance.

Lastly, the Hybrid typology represents a novel approach proposed in this study, integrating the salient features of the aforementioned typologies. This integrative method is envisioned to leverage the combined attributes of Traditional Neural Networks, Co-attention mechanisms, and Transformers, aiming to explore potential synergies and subsequently augment the performance in VQA tasks. The amalgamation of the characteristics of the distinct types is intended to foster a comprehensive understanding and enhanced interpretative capability in VQA applications.

Further detailed expositions of the stated methodologies and their individual characteristics, components, and implementations are provided in the ensuing sections of this document, elucidating the intricate mechanisms and theoretical underpinnings of each approach.
\subsubsection{Traditional neural network approach}
\sstitle{LSTM-Q \citep{DBLP:conf/iccv/AntolALMBZP15}}
The LSTM-Q model, interestingly, delivers commendable results on VQA datasets \citep{DBLP:conf/iccv/AntolALMBZP15} and CLEVR \citep{Johnson_2017_CVPR} even in the absence of image-based input. This model interprets the question through acquired word embeddings and subsequently employs a word-level LSTM \citep{10.1162/neco.1997.9.8.1735} for processing. The concluding hidden state of the LSTM is channeled into a multi-layer perceptron (MLP), which then estimates a probabilistic distribution of potential answers. Given its exclusive dependence on question data, the model can only accommodate biases that are conditional on the question.

\sstitle{CNN+LSTM \citep{Johnson_2017_CVPR}} The model accompanied by the dataset utilizes a combination of CNN-based image embedding and LSTM-based question embedding. The embeddings obtained from these two components are merged using point-wise multiplication, and the resulting embeddings are subsequently fed into a multi-layer perceptron classifier to predict the probability distribution of the answer. 
\subsubsection{Co-attention approach}
\sstitle{ViHieCoAtt \citep{tran-etal-2021-vivqa}}
The Alternating Co-attention mechanism functions through an iterative procedure, concentrating selectively on either the question or image features, guided reciprocally by the attributes of the image or question. This recurrent operation permits dynamic and adjustable allocation of attention, aiding in the amalgamation of pertinent information from both modalities.
The ViHieCoAtt approach employed PhoW2V embeddings \citep{tuan-nguyen-etal-2020-pilot}, generated through pre-training both 100-dimensional and 300-dimensional syllable embeddings alongside 100-dimensional and 300-dimensional word embeddings, using the Word2Vec skip-gram model \citep{DBLP:journals/corr/abs-1301-3781}. This initial training phase was undertaken on expansive Vietnamese text corpora, encompassing both syllable and word levels, amounting to 20GB in total.

\sstitle{ViMCAN \citep{NGUYEN2023101868}} ViMCAN stands as the Vietnamese iteration of Modular Co-attention Networks \citep{8953581}. MCAN employs a layer comprised of MCA. The MCA layer represents a modular amalgamation of two fundamental attention units: the self-attention (SA) unit and the guided-attention (GA) unit, which draw inspiration from the scaled dot-product attention mechanism. MCAN endeavors to concurrently investigate inter-modality and intra-modality relations, yielding commendable outcomes. It introduces a profound Modular Co-Attention Network constructed of Modular Co-Attention (MCA) layers organized in a cascaded manner in depth, allowing for the exploration of nuanced attention dynamics and interactions within and across modalities.
\subsubsection{Transformer approach}
\sstitle{BARTPhoBEiT \citep{tran2023bartphobeit}} BARTPhoBEiT is a our previous novel integration of the BARTPho \citep{bartpho2022} and BEiT-3 \citep{wang2023image} models, specifically tailored for the Vietnamese language. This innovative model incorporates pre-trained Sequence-to-Sequence and bidirectional encoder representations derived from Image Transformers. The BARTPhoBEiT model's performance is comprehensively evaluated using the ViVQA datasets \citep{tran-etal-2021-vivqa}, achieving state-of-the-art (SOTA) results. This evaluation offers valuable insights into the model's effectiveness and suitability for Visual Question Answering (VQA) tasks in the context of the Vietnamese language. In this paper, we extend its capabilities to handle visual reasoning tasks with some minor improvements. 
\subsubsection{Hybrid approach}
\sstitle{PhoViT} In Section \ref{sec:proposedmethod}, we delineate the conceptualization and construction of PhoViT, an innovative model we have developed, embodying a synergy of neural network methodologies, co-attention mechanisms, and a framework grounded in transformer-based methodologies. The essence of our approach is the utilization of PhoW2V, a specialized construct of a neural network, designed for the embedding of questions, and the employment of Vision Transformer (ViT) for the intricate embedding of images.
Subsequently, our approach incorporates a sophisticated multimodal fusion methodology \citep{ZHANG20211}, leveraging stacking attention—commonly referred to as co-attention, to effectually amalgamate informational constituents derived from both images and questions. This meticulous integration is pivotal, enabling the coherent synthesis of multiform information modalities, and contributing to the efficacious convergence of visual and textual elements.
This novel approach exhibits substantial promise, elucidating its potential ramifications through systematic experimental investigations within the disciplinary contexts of Visual Question Answering (VQA) and Visual Reasoning tasks. 
\subsection{Evaluation metrics}
Consistent with the preceding research conducted by \citep{tran-etal-2021-vivqa, nguyen2023vlsp, Johnson_2017_CVPR, hudson2019gqa}, elucidating the evaluation metrics utilized for gauging the model's efficacy is crucial before delving into the analysis of the experimental outcomes. The appraisal in this research encompasses four pivotal performance metrics: F1 score, Precision, Recall, and Accuracy. The derivation of the F1 score and Accuracy for each individual response is achieved through the tokenized forms of the anticipated answer (AA) and the standard answer (SA). Subsequently, to calculate the cumulative F1 score ($F1_{Overall}$), the F1 scores corresponding to all queries within a specific subset are averaged.
\begin{equation}
     Precision (P) = \frac{SA \cap AA}{AA}
\end{equation}
\begin{equation}
      Recall (R) = \frac{SA \cap AA}{SA}
\end{equation}
\begin{equation}
      F1 = \frac{2 \times P \times R }{P+R}
\end{equation}
\begin{equation}
      F1_{Overall} =  \frac{1}{N} \sum_{i=1}^{N} F1_i
\end{equation}
\begin{equation}
       Accuracy = \frac{AA}{SA}
\end{equation}
Drawing inspiration from the work of Nguyen et al. \citep{NGUYEN2023101868}, this study incorporates several evaluative metrics, notably (BLEU) \citep{papineni-etal-2002-bleu}, ROGUE-L \citep{DBLP:journals/corr/abs-1803-01937}, and METEOR \citep{banerjee-lavie-2005-meteor}. These metrics have been meticulously selected to ensure rigorous and comprehensive assessment methodologies in our experimental framework.

\sstitle{BLEU \citep{papineni-etal-2002-bleu}} BLEU (Bilingual Evaluation Understudy) is a widely acclaimed metric in the realm of natural language processing and computational linguistics, employed to assess the correspondence between machine-generated text and a set of reference texts. This metric is crucial as it quantitatively evaluates the coherency, relevance, and alignment of the generated text in comparison to the reference, providing an objective measure of the model's performance in generating linguistically and contextually accurate text.

In the realm of measurement techniques, the predominant emphasis of this metric is directed towards the hue characteristics intrinsic to the precision measurement. The BLEU metric was rooted in a pair of pivotal observations: first, the frequency of n-gram entities within a hypothesis (hypo) ought not to surpass its manifestation within the reference (ref). Second, any hypo with a magnitude exceeding that of a corresponding ref should be subjected to a diminutive weighting factor (designated as a penalty weight). In more precise terms, the scoring system for a hypo, given its associated ref, can be mathematically represented as:
\begin{equation}
   \text{score}_{\text{token}} = \frac{\text{Count}_{\text{clip}}(\text{token})}{\text{Count}(\text{token})}  
\end{equation}

Subsequently, utilizing this equation, the cumulative scoring methodology for all hypo entities within a dataset is expressed as:
\begin{equation}
    p_n = \frac{\sum_{h \in hypothesis} \sum_{token \in h} \text{Count}_{\text{clip}}(token)}{\sum_{h \in hypothesis} \sum_{token \in h} \text{Count}(token)} 
\end{equation} 
Herein, \( n \) signifies the integer value associated with the selected n-gram.

While the equation for \( p_n \) inherently addresses scenarios where the magnitude of hypo exceeds one of ref, there remains the contingency wherein the hypo's length is inferior to its ref counterpart. Addressing this, if \( c \) represents the collective length of all hypos within the dataset and \( r \) denotes the total length of all refs in the dataset, then the penalty weight, catering to hypos of length surpassing that of refs, is formulated as:
\begin{equation}
     BP = e^{1-\frac{r}{c}} 
\end{equation}

It is self-evident that \( BP \) equates to 1 when \( c > r \).

Conclusively, the resultant BLEU score is extrapolated through the equation:
\begin{equation}
    \log BLEU = \min(1 - \frac{r}{c}, 0) + \sum_{n=1}^{N} w_n \log p_n 
\end{equation}

\sstitle{ROGUE \citep{DBLP:journals/corr/abs-1803-01937}} ROUGE (Recall-Oriented Understudy for Gisting Evaluation) is a set of evaluation metrics that are commonly used to measure the quality of summaries by comparing them to reference summaries. Among the several variants of the ROUGE metric, ROUGE-L (Longest Common Subsequence) is one of the most frequently employed. 

Ganesan et al \citep{DBLP:journals/corr/abs-1803-01937} delineated the measures of recall, denoted as \( R \), and precision, represented as \( P \), employing the Longest Common Subsequence (LCS) between two entities: hypo and ref. The recall, denoted as \( R_{LCS} \), is formulated as the ratio of the LCS of hypo and ref to \( m \), which signifies the length of ref, as depicted in the equation:
\begin{equation}
 R_{LCS} = \frac{LCS(hypo, ref)}{m} 
\end{equation}

Conversely, the precision, represented as \( P_{LCS} \), is given by the proportion of the LCS of hypo and ref to \( n \), where \( n \) characterizes the length of hypo, expressed as:
\begin{equation}
   P_{LCS} = \frac{LCS(hypo, ref)}{n} 
\end{equation}

Subsequently, the metric ROUGE-L is articulated as a combination of the aforementioned recall and precision values, encapsulated in the relation:
\begin{equation}
    ROUGE = \frac{(1 + \beta^2) R_{LCS} P_{LCS}}{R_{LCS} + \beta^2 P_{LCS}} 
\end{equation}

\sstitle{METEOR \citep{banerjee-lavie-2005-meteor}}
In the realm of machine translation evaluation, both BLEU and ROUGE employ n-gram tokens as their foundational components for token definition. Contrarily, METEOR (Metric for Evaluation of Translation with Explicit ORdering) posits that there exist instances wherein interchanging the positions of n-gram tokens doesn't necessarily alter the overall semantic essence of the sentence. Nonetheless, such configurations tend to be penalized with diminished scores under the BLEU and ROUGE metrics. Addressing this predicament, Banerjee et al. introduced the notion of alignment between the hypothesis (hypo) and the reference (ref). In this context, alignments are construed as a collection of mappings, with each mapping representing a distinct association between tokens present in both the hypo and the ref. It's pivotal to underscore that, within this framework, a token is characterized as a 1-gram entity.

It considers various linguistic phenomena, including word-to-word matches, stemming, synonymy, paraphrasing, and word order. The METEOR score is computed using precision, recall, and a harmonic mean of these with a penalty factor for word order differences.

For any given hypothesis (hypo) and reference (ref), numerous alignments might be discerned.  With this context established, the precision \(P\) derived from the hypo and ref based on their corresponding alignment is articulated as: 
\begin{equation}
     P = \frac{m}{wh} 
\end{equation}

Meanwhile, the recall \(R\) in relation to the hypo and ref is formulated as:
\begin{equation}
R = \frac{m}{wr} 
\end{equation}

Subsequently, their correlation is expressed by the F-measure:
\begin{equation}
    F_{mean} = \frac{10PR}{R + 9P} 
\end{equation}

Drawing parallels to the BLEU metric, METEOR outlines a penalty factor for a hypo that either surpasses or falls short of its corresponding ref, factoring in the mutual tokens present in both hypo and ref. In this regard, the penalty weight \(p\) is delineated as:
\begin{equation}
    p = 0.5 \times \left(\frac{c}{um}\right)^3 
\end{equation}

Here, \(c\) denotes the count of shared unigrams between the hypo and its ref, while \(um\) signifies the cumulative unigrams observed across both the hypo and ref. Integrating the penalty weight with the correlation between precision \(P\) and recall \(R\), the METEOR score, with respect to the hypo and its ref, is defined as:
\begin{equation}
    M = F_{mean} \times (1 - p)
\end{equation}

\subsection{Experimental Results}


        
       
\begin{table*}[!h]
\caption{The outcomes of the experimentation reflect the performances of diverse Vietnamese VQA systems evaluated on the ViCLEVR dataset.}
\centering
\begin{tabular}{llllllll}
\hline
\multicolumn{1}{l}{Model}                              & \multicolumn{1}{l}{Accuracy} & \multicolumn{1}{l}{Precision} & \multicolumn{1}{l} {Recall} &  \multicolumn{1}{l}{F1$_{overall}$}  & \multicolumn{1}{l}{BLEU} & \multicolumn{1}{l}{ROGUE} &METEOR \\ \hline
\multicolumn{8}{c}{Traditional Neural Network}                                                                                                    \\ \hline
\multicolumn{1}{l}{LSTM-Q \citep{DBLP:conf/iccv/AntolALMBZP15}} & \multicolumn{1}{l}{0.207}                                       & \multicolumn{1}{l}{0.008}                               & \multicolumn{1}{l}{0.038}                            & \multicolumn{1}{l}{0.013} & \multicolumn{1}{l}{0.199} &  \multicolumn{1}{l}{0.212} &  0.0981                              \\
\multicolumn{1}{l}{CNN+LSTM \citep{Johnson_2017_CVPR}}        & \multicolumn{1}{l}{0.217}                                       & \multicolumn{1}{l}{0.0166}                              & \multicolumn{1}{l}{0.041}                            & \multicolumn{1}{l} {0.024} &  \multicolumn{1}{l} {0.258} &    \multicolumn{1}{l} {0.241} &   0.112                          \\ \hline
\multicolumn{8}{c}{Co-attention}                                                                                                                                                                                                                                                                          \\ \hline
\multicolumn{1}{l}{ViHieCoAtt \citep{tran-etal-2021-vivqa}}     & \multicolumn{1}{l}{0.291}                                       & \multicolumn{1}{l}{0.091}                               & \multicolumn{1}{l}{0.100}                            & \multicolumn{1}{l}{0.083} & \multicolumn{1}{l}{0.349}  & \multicolumn{1}{l}{0.326} &   0.164                             \\ 
\multicolumn{1}{l}{ViMCAN \citep{NGUYEN2023101868}}             & \multicolumn{1}{l}{0.445}                                       & \multicolumn{1}{l}{0.315}                               & \multicolumn{1}{l}{0.304}                            & \multicolumn{1}{l}{0.281}   & \multicolumn{1}{l}{0.500} & \multicolumn{1}{l}{0.478} & 0.255

\\ \hline

\multicolumn{8}{c}{Transformer}                                                                                                                                                                                                                                                   \\ \hline
\multicolumn{1}{l}{BARTPhoBEiT  \citep{tran2023bartphobeit}}    & \multicolumn{1}{l}{0.466}                                       & \multicolumn{1}{l}{0.387}                               & \multicolumn{1}{l}{0.325}                            & \multicolumn{1}{l}{0.308}           & \multicolumn{1}{l}{0.513} & \multicolumn{1}{l}{0.491} &     0.263                 \\ \hline

\multicolumn{8}{c}{Hybrid}                                                                                               \\ \hline
\multicolumn{1}{l}{\textbf{PhoViT (Ours)}}                      & \multicolumn{1}{l}{\textbf{0.468}}             & \multicolumn{1}{l}{\textbf{0.332}}    & \multicolumn{1}{l}{\textbf{0.324}}  & \multicolumn{1}{l}{\textbf{0.316}} & \multicolumn{1}{l}{\textbf{0.514    } } & \multicolumn{1}{l}{\textbf{0.494 }    } &  \textbf{0.264 } \\
\hline
\end{tabular}
\label{tab:ex}
\end{table*}
The results presented in \autoref{tab:ex} showcase that our proposed model surpasses all baseline models in performance. The baselines consist of two models: a "blind" LSTM model that only has access to the questions, and an LSTM+CNN model employing a neural network approach. These baselines achieve relatively low-performance results, ranging from 20\% to 21.7\%. Specifically, the LSTM model exhibits a success rate of only 20.7\% for open query questions and performs only slightly above chance for binary question types. 

Inclusive of the baseline models, our evaluation encompasses two contemporary models, ViHieCoAtt and ViMCan, which leverage Co-attention mechanisms and exhibit proficiency on the ViVQA dataset. Furthermore, we present a novel model, denoted as PhoViT, which constitutes a pioneering visual reasoning approach tailored specifically for the Vietnamese language. Detailed elaboration on PhoViT can be found in  \autoref{sec:proposedmethod}. In order to discern the underlying factors contributing to the superior performance of our proposed model compared to others, a comprehensive analysis will be conducted through a series of ablation studies. These studies will draw upon the findings presented in \autoref{sec:QuestionCategory}, \autoref{sec:QuestionType}, \autoref{sec:QuestionSize} of the preceding subsections as a foundation for further investigation.
\subsection{Analysis by question category}
\label{sec:QuestionCategory}
\begin{figure*}[!h]
    \centering
    \includegraphics[width=1\linewidth]{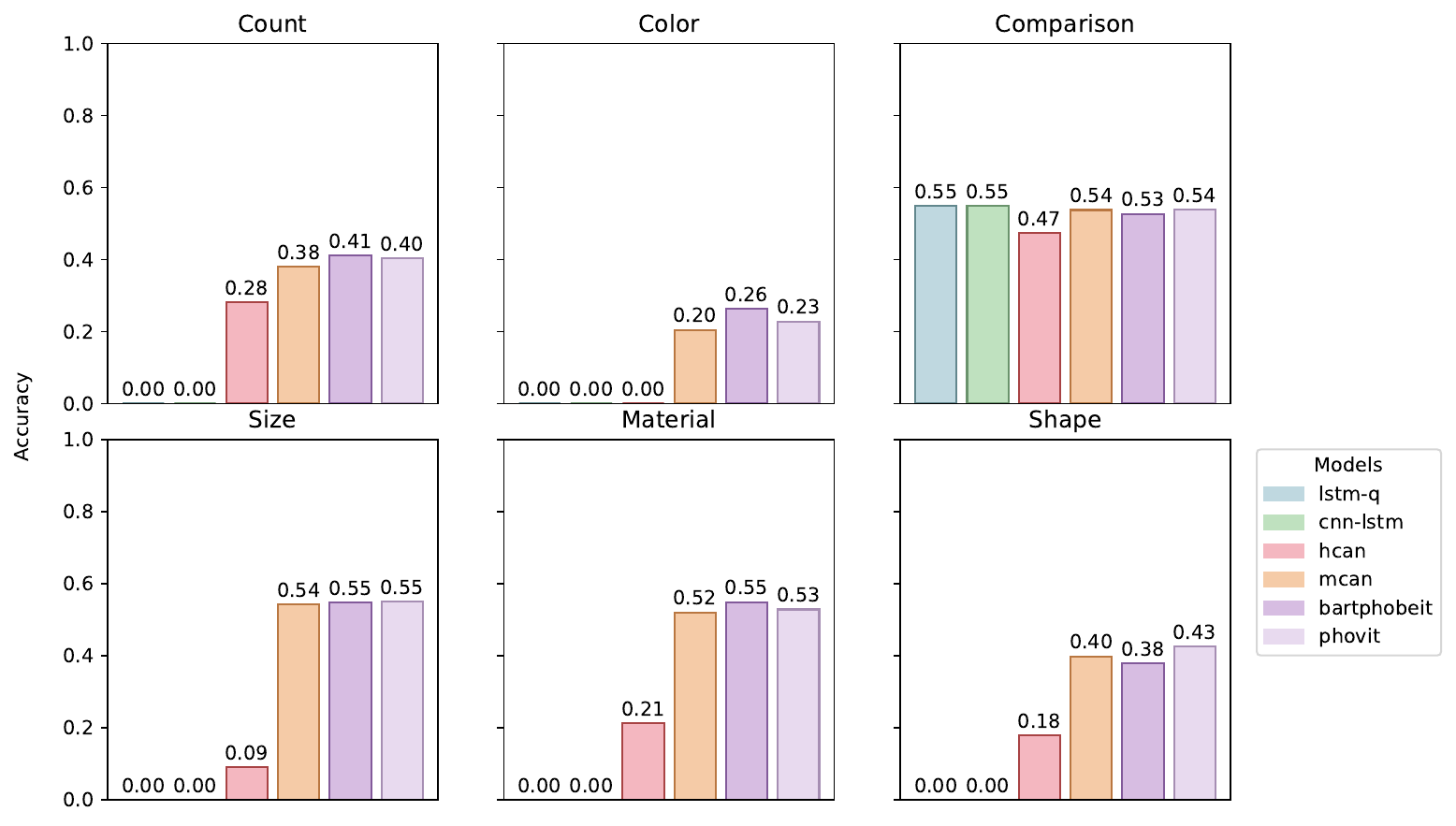}
    \caption{The performance with respect to accuracy across distinct question categories is evaluated for six distinct VQA techniques, two of which are original contributions presented herein. The evaluation is conducted on the ViCLEVR dataset, with superior values indicating enhanced performance}
    \label{fig:questionCategory}
\end{figure*}

We can employ the programmatic representation of questions to assess the model's performance across various forms of reasoning. Initially, we assess the model's proficiency in handling each distinct question type, delineated by the outermost function in the program. The results are visualized in \autoref{fig:questionCategory}, and a more in-depth analysis of these outcomes will be provided in the subsequent discussion.

\sstitle{Count} Counting questions inquire about the quantity of objects that meet specific criteria (e.g., \vn "Có bao nhiêu vật trụ màu xanh?" \en ("How many blue cylinders are there?")).  Given that images consist of varying numbers of objects ranging from three to ten, and counting questions pertain to subsets of these objects, achieving a uniform distribution of answers is a formidable challenge. In the context of counting questions, both the "blind" LSTM and the LSTM+CNN models achieve accuracies that are nearly 0\%. This outcome arises from the fact that these models primarily produce responses indicating values of "Yes" and "No". This pattern of results suggests that the dataset exhibits minimal question-conditional bias for counting-related inquiries. Our rejection sampler approach strives to encourage a more uniform answer distribution for these questions without imposing it as a strict requirement. Consequently, this approach introduces a bias that is contingent on the nature of the question, as evidenced by the accuracy rates of 28\% and 38\% achieved by ViHieCoAtt and ViMCAN.
Intriguingly, our two new proposed models, the BARTPhoBEiT and PhoViT outperform comparably to existing ones, indicating that the Transformer features contribute limited information relevant to the counting task. The BARTPhoBEiT model exhibits a slight improvement in performance, but its absolute accuracy remains modest at 41\%.

\sstitle{Color} Questions related to color seek to acquire information regarding the specific object's chromatic characteristics (e.g., \vn "Màu sắc của khối lập phương phía trên vật tròn màu xanh da trời là gì?" \en ("What is the color of the cubic object above the sky-blue circular one?")). The LSTM, LSTM+CNN, and VieHieCoAtt models exhibit a notable decline in performance, yielding an approximate accuracy of 0\% when confronted with this specific question type. This observation can be attributed to the inherent complexity of these questions within our dataset, often intertwined with other question types, necessitating a more intricate chain of logical reasoning for accurate prediction.

In contrast, our BARTPhoBEiT and PhoViT models consistently outperform other methodologies. This superior performance can be attributed to the image embedding capabilities of these models, which rely on the Vision Transformer (ViT) architecture, thereby enhancing their capacity to extract and discern object colors with superior precision compared to alternative techniques. However, despite this notable achievement, the overall accuracy of these models remains marginally below the 25\% threshold, underscoring the formidable challenge presented by our dataset, even when employing state-of-the-art Transformer-based approaches.

\sstitle{Comparison} Comparison queries seek to ascertain whether two entities possess an equivalent quantitative magnitude with regard to a particular characteristic (e.g., \vn "Có phải vật hình trụ lớn hơn vật màu xanh không?" \en ("Is the cylinder larger than the blue object?")). The exclusively acceptable responses to such inquiries are confined to the binary options of "yes" and "no." In the domain of this particular question type, LSTM-based models exhibit superior performance compared to alternative techniques, primarily owing to the inherent strengths of LSTM models in providing precise "Yes" and "No" responses. Nevertheless, our proposed methodologies yield results slightly lower, with only a 1\% discrepancy. This outcome serves as a testament to the general competence of our proposed model in effectively addressing this specific question category.

\sstitle{Size} Questions regarding size focus on ascertaining whether an object possesses greater or lesser dimensions (e.g., \vn "Có một khối trụ phía trên vật hình vuông bên trái vật hình tròn, kích thước của nó như thế nào?" \en ("There is a cylindrical object above the square-shaped object to the left of the circular object; what are its size?")). Within the category of this particular question type, our BARTPhoBEiT and PhoViT models exhibit superior performance, achieving an impressive accuracy rate of nearly 55\%. It's worth noting that questions of this type in our dataset often involve a blend of attributes, such as "material" and "color," necessitating a more intricate chain of reasoning compared to questions in other datasets. Consequently, the transformer-based techniques manifest their effectiveness in addressing these multifaceted questions, surpassing traditional approaches.

Furthermore, the utilization of a multimodal fusion layer in conjunction with deep stacked and multiway transformers underscores its effectiveness in enhancing the overall performance and accuracy of our models in handling such complex question types.

\sstitle{Material} Material inquiries seek information pertaining to the substance or composition of an object, such as whether it is constructed from specific materials (e.g., \vn "Chất liệu của vật màu xanh lớn kế bên vật trụ màu vàng là gì?" \en ("What is the material of the large green object next to the yellow cylinder?")). The outcomes for this specific question type closely resemble those of questions related to size. In this category, our proposed models consistently attain state-of-the-art performance. ViMCAN, while displaying commendable results in this context, owes its success to the integration of a deep fusion layer for answer prediction. Nevertheless, it falls short of surpassing our results due to the distinguishing factor lying in the classifier layer. Notably, our models employ a vision-language Feed-Forward Network (FFN), a method superior to the traditional FFN, thereby yielding more favorable outcomes.

\sstitle{Shape} Shape question inquiries into the geometric configuration of an object, such as whether it conforms to specific shapes (e.g., \vn "Hình dạng của vật kim loại màu tím nhỏ kế bên vật tròn màu xanh lá cây là gì?" \en ("What is the shape of the small purple metal object next to the green circular one?")). Within this particular question category, our PhoViT model demonstrates a significant and notable advantage over all other models. This enhanced performance can be attributed to the innovative fusion of an attention mechanism with the Transformer-based technique, which empowers PhoViT to excel in addressing these questions effectively.

It is worth noting that other models that incorporate attention mechanisms also exhibit improved performance in this context. This improvement can be attributed to the attention mechanism's inherent capacity to facilitate a focused examination of the target object, thereby aiding in the identification of its specific shape. However, our PhoViT model, with its unique combination of attention mechanisms and Transformer-based techniques, stands out as the top-performing solution in this question category.

\subsection{Analysis by linguistic question type}
\label{sec:QuestionType}
\begin{figure*}[!h]
    \centering
    \includegraphics[width=1\linewidth]{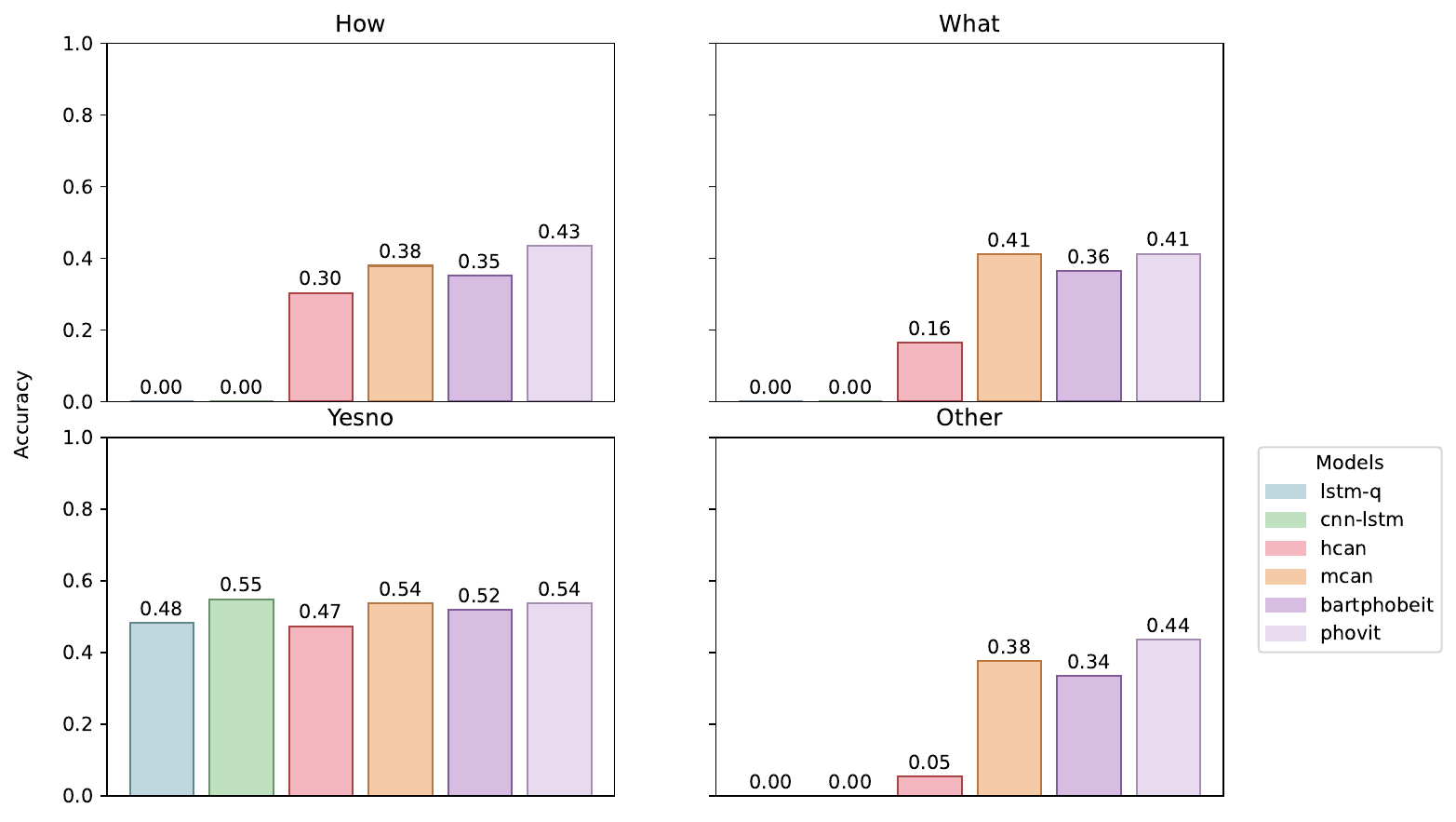}
    \caption{The performance with respect to accuracy across distinct question type is evaluated for six distinct VQA techniques, two of which are original contributions presented herein. The evaluation is conducted on the ViCLEVR dataset, with superior values indicating enhanced performance}
    \label{fig:questionLinCategory}
\end{figure*}

In the context of our dataset, the data can be categorized into four primary types: "What," "How," "Yes/No," and "Other," driven by the inherent characteristics of the associated images and data. In the course of this experimentation, we embark on a comprehensive research analysis using six distinct VQA methodologies applied to these categories. The objective is to gain a more profound comprehension of the implications and impacts inherent in this categorization. \autoref{fig:questionLinCategory} illustrates our investigation into this particular question category. 

Within our dataset, questions beginning with "what" exhibit a noteworthy diversity in terms of question phrasing and tend to be longer in length, resulting in heightened complexity for these question types.
Similarly, questions commencing with "How" pose a considerable challenge, necessitating the model's ability to comprehensively grasp the textual context in order to furnish precise responses. This category encompasses variations such as "how many" and "how much," which introduce additional intricacies in the Vietnamese language due to the diverse linguistic structures employed.
Conversely, "Yes/No" questions inherently offer limited information for answer retrieval, and as a consequence, most models demonstrate adept performance when tackling queries of this nature.
\subsection{Analysis by question size}
\label{sec:QuestionSize}
\begin{figure*}[!h]
\begin{minipage}{0.45\linewidth}
  \centering
    \begin{tabular}{cc}
    \hline
        \textbf{Group} & \textbf{Length (n)}  \\ \hline
        Short & $n \leq Q1 (16.0)$  \\ 
        Medium  & $Q1 (16.0) <n \leq Q2 (19.0)$ \\ 
        Long  & $Q2 (19.0) <n \leq Q3 (24.0)$  \\ 
        Very long  & $n> Q3 (24.0)$\\ 
        \hline
    \end{tabular}
  \label{tab:tablen}
\end{minipage}
\begin{minipage}{0.45\linewidth}
  \centering
  \includegraphics[width=\linewidth]{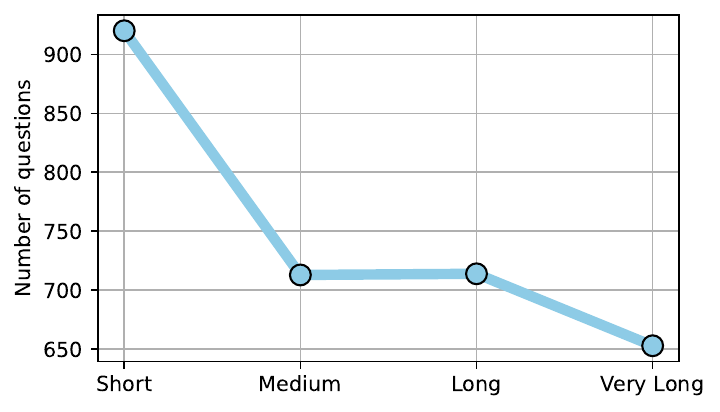}

\end{minipage}
\hfill
\caption{Categorization of questions based on their total word count in test set. In our dataset, we partition it into four distinct length categories, each aligning with a specific quartile range: Q1, Q2, Q3, and Q4, representing the 25th, 50th, 75th, and 100th percentiles, respectively. These quartile ranges correspond to question lengths of 16, 19, 24, and greater than 24 words.}
\label{fig:QuesSizeDecrip}
\end{figure*}

\begin{figure*}[!h]
    \centering
    \includegraphics[width=0.8\linewidth]{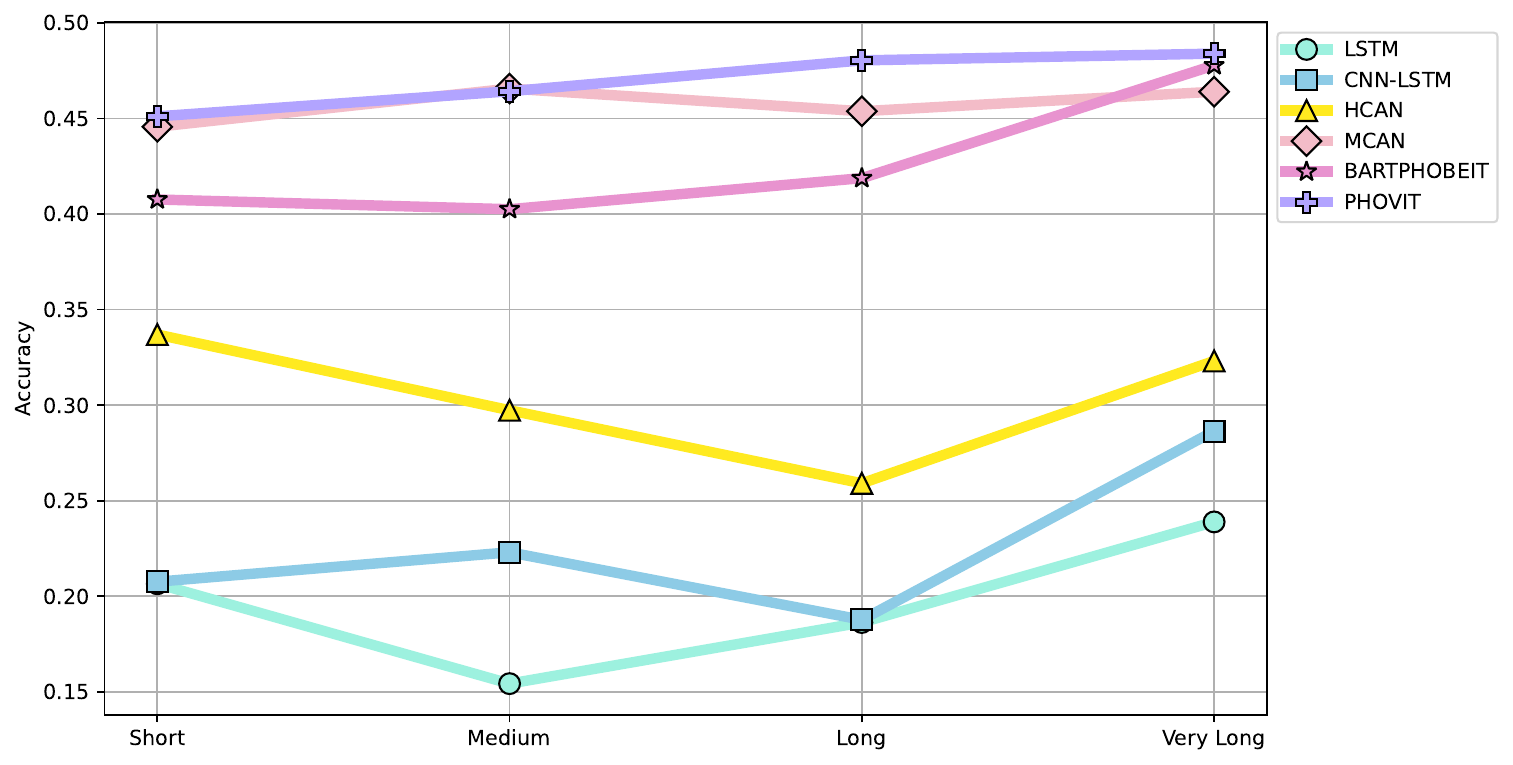}
    \caption{Accuracy analysis across question lengths in various VQA Systems, including two proposed models}
    \label{fig:questionSize}
\end{figure*}

For the sake of facilitating a comprehensive analysis, we partition questions into distinct categories determined by their length, specifically the overall count of words they encompass. To be precise, the categorization of questions is delineated in \autoref{fig:QuesSizeDecrip}. Subsequently, we subject both baseline models and our novel proposed methods to evaluation within each respective question group, thereby extending this assessment to encompass both questions and their corresponding answers as per this categorization. In its initial phase, our investigation entails a comprehensive examination of the outcomes achieved by the baseline models and the proposed methodologies across the aforementioned question groups. The particulars of this evaluation are expounded upon in \autoref{fig:questionSize}.

From a conceptual perspective, it is reasonable to anticipate that longer inquiries would entail higher levels of complexity, as they inherently involve a more extensive sequence of logical steps. Nevertheless, it is of significance to underscore that a substantial proportion of questions exhibit a noteworthy ability to yield accurate responses, even in instances where specific subtasks remain unresolved. This phenomenon finds illustration in Figure \ref{fig:questionSize}, where the successful resolution of the posed question does not hinge on a precise identification of the particular large blue cylinder. This observation is grounded in the established understanding that objects prominently positioned to the left of a cylinder inherently belong to the category of cylinders.

Notably, the question depicted in Figure \ref{fig:questionSize} defies classification as degenerate, given that its entire construct plays a vital role in disambiguating references to distinct objects—namely, two blue cylinders and two rubber cylinders. Despite this intricate nature, the question manifests a relatively compact effective size due to its capacity to be addressed with precision without the necessity to fully unravel the mentioned references.
A discernible trend emerges, wherein the error rate exhibited by all models exhibits an upward trajectory concurrent with an expansion in question size. This discernible pattern underscores the challenges encountered by models when grappling with intricate reasoning pathways intrinsic to more extended questions.




\begin{figure*}[!h]
    \centering
    \includegraphics[width=1\linewidth]{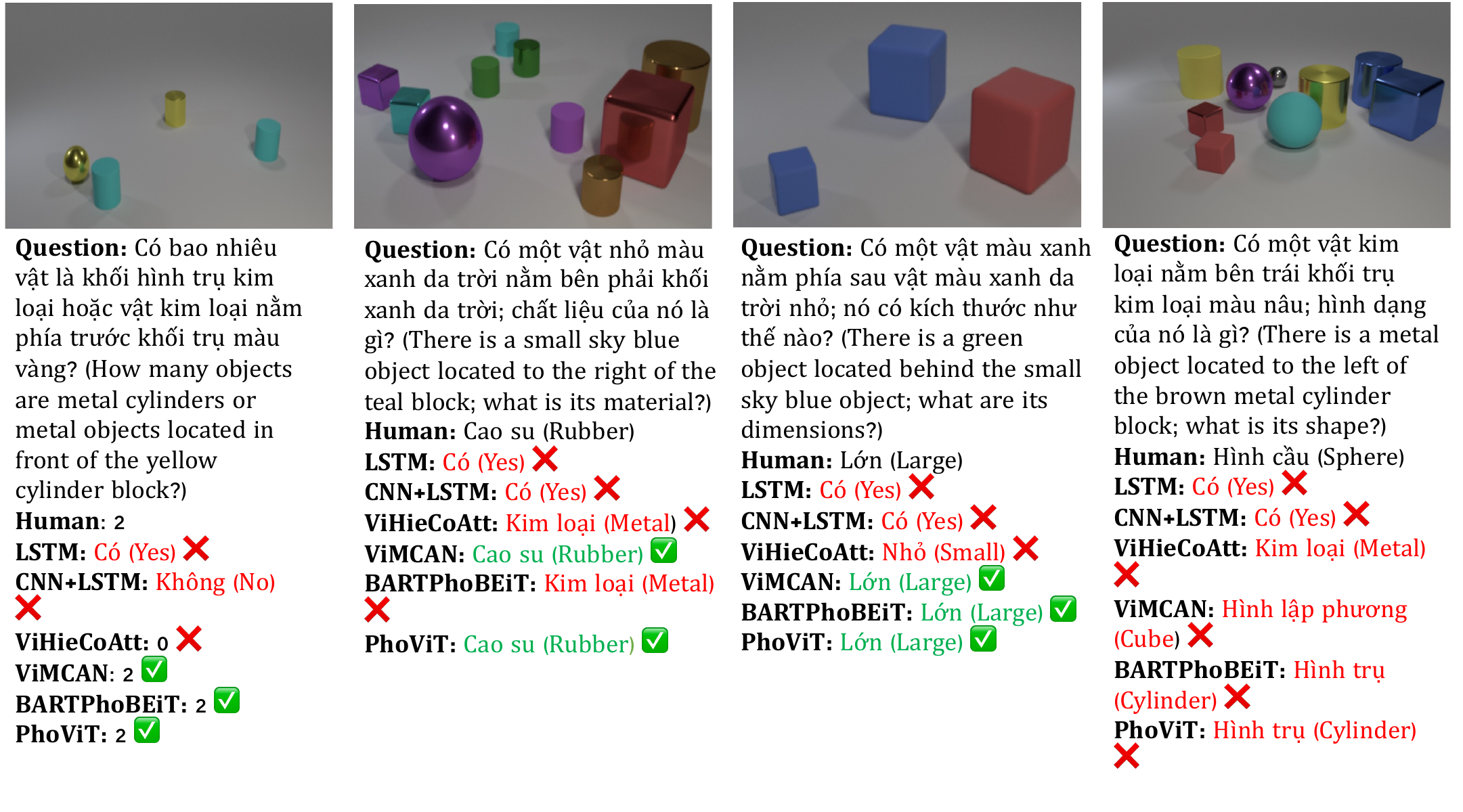}
    \caption{Several examples of our two proposed methods and baselines.}
    \label{fig:answerExample}
\end{figure*}

%% file: tex/6-Discussion.tex

%% file: tex/7-Conclusion.tex
\section{Conclusion and future work}
\label{sec:conclude}
In this paper, we introduced the ViCLEVR dataset, designed for visual reasoning and visual question answering in Vietnamese. The dataset generation process was elaborated, and baseline experiments were conducted, along with the introduction of new measures to gain deeper insights into the behavior and performance of models. We believe that this benchmark will serve as a valuable resource for advancing research in Vietnamese VQA by promoting more integrated approaches that intertwine visual reasoning and visual question answering, two thriving fields that have often been explored independently.

Moreover, we presented PhoViT, a new hybrid multimodal model, and demonstrated its robust performance on multimodal reasoning tasks using the proposed dataset, showcasing its potential in the Vietnamese language. We firmly believe that the ViCLEVR dataset and PhoViT will inspire and facilitate the development of more compositional, interpretable, and effective reasoning models, thus propelling research in Vietnamese scene understanding and visual question answering to new heights.

Furthermore, it is worth noting that the ViCLEVR dataset, despite being the inaugural visual reasoning dataset in the Vietnamese language, is relatively modest in size when juxtaposed with its English-language visual reasoning counterparts. In our forthcoming research endeavors, we intend to augment the ViCLEVR dataset by incorporating a larger volume of images and question-answer pairs. Additionally, we plan to enrich the dataset by introducing a more diverse set of answers and increasing the answer length for questions. This expansion will enable a more comprehensive evaluation of visual reasoning responses.
Moreover, we have strategic plans to extend the ViCLEVR dataset into a multilingual, low-resource language Visual Question Answering (VQA) dataset. This initiative is aimed at providing a valuable resource for research in multilingual VQA, encompassing the Vietnamese language as well. 